\pdfoutput=1
\documentclass[11pt]{article}

\usepackage[preprint]{acl}
\makeatletter
\IfFormatAtLeastTF{2025-06-01}{\@ifpackagelater{lineno}{2026/01/01}{%
  \RemoveFromHook{build/column/before}[lineno]%
  \AddToHook{build/column/before}{%
    \protected@write\@auxout{}{%
      \string\def\string\@LN@column{\if@firstcolumn2\else1\fi}}}%
}{}}{}
\makeatother

\usepackage{times}
\usepackage{latexsym}

\usepackage[T1]{fontenc}

\usepackage[utf8]{inputenc}

\usepackage{microtype}

\usepackage{inconsolata}

\usepackage{graphicx}
\usepackage{amsmath,amssymb}
\usepackage{booktabs}
\usepackage{tabularx}
\newcolumntype{Y}{>{\centering\arraybackslash}X} %
\usepackage{colortbl}
\usepackage[ruled,vlined]{algorithm2e}
\usepackage{pgfplots}
\pgfplotsset{compat=1.18}

\newcommand{\proj}{HR-Merge}
\definecolor{oursblue}{RGB}{235,244,252}
\newcommand{\observationbox}[2]{%
  \noindent\fbox{%
    \parbox{\dimexpr\columnwidth-2\fboxsep-2\fboxrule\relax}{%
      \textbf{Observation #1.} \textit{#2}%
    }%
  }\par\vspace{0.6em}%
}
\newtheorem{proposition}{Proposition}
\newtheorem{theorem}{Theorem}
\newtheorem{assumption}{Assumption}

\title{HR-Merge: Model Merging with Headroom-Weighted Residual Refinement}

\author{%
  \textbf{Jiakang Huang\textsuperscript{1}}\thanks{Equal contribution.}\hspace{1.0em}\textbf{Xueyan Zhang\textsuperscript{2}}\footnotemark[1]\hspace{1.0em}\textbf{Renyi Cai\textsuperscript{2}}
\\[3pt]
  \textbf{Jingcheng Niu\textsuperscript{3}}\hspace{1.0em}\textbf{Jiaru Li\textsuperscript{4}}\hspace{1.0em}\textbf{Yining Wang\textsuperscript{5}}\hspace{1.0em}\textbf{Josh Zhang\textsuperscript{6}}\hspace{1.0em}\textbf{Jiahe Jiang\textsuperscript{1}}
\\[3pt]
  \textbf{Jinman Zhao\textsuperscript{5}}\thanks{Corresponding author: \href{mailto:jzhao@cs.toronto.edu}{jzhao@cs.toronto.edu}}\hspace{1.0em}\textbf{Gerald Penn\textsuperscript{5}}\hspace{1.0em}\textbf{Baharan Mirzasoleiman\textsuperscript{7}}
\\[8pt]
  \small\mdseries \textsuperscript{1}University of British Columbia\hspace{1.1em}\textsuperscript{2}Independent Researcher\hspace{1.1em}\textsuperscript{3}Technische Universit\"at Darmstadt
\\[1pt]
  \small\mdseries \textsuperscript{4}Northwestern University\hspace{1.1em}\textsuperscript{5}University of Toronto\hspace{1.1em}\textsuperscript{6}Token Canopy\hspace{1.1em}\textsuperscript{7}University of California, Los Angeles
\\[1pt]
  \small\mdseries\texttt{jhuang74@student.ubc.ca, xueyanzhang27@gmail.com}
}

\begin{document}
\maketitle
\begin{abstract}
Model merging combines experts fine-tuned from a shared backbone, such as
LoRA adapters of a large language model, into one multi-task model without
further training. Existing methods apply one rule to every task, yet the
accuracy they lose is uneven across tasks. We show that the lost accuracy
remains recoverable from each task's residual, its task vector minus the merged update, and that a task's validation headroom, the share of its
expert validation accuracy the merge loses, indicates where recovery
pays off. Residuals from different tasks interfere, so they must be merged
rather than summed. We propose \proj{}, a training-free wrapper around any
merging method. It weights each residual by its task's headroom, re-merges
the weighted residuals with the base operator, and adds the result back at a
strength selected on validation, falling back to the base merge when no
strength helps. On six NLI adapters of LLaMA-3 8B, \proj{} lifts TIES
from $93.29$ to $94.29$ normalized accuracy, above all tested bases, mainly by
recovering its weakest task. In vision, on three CLIP backbones,
up to 20 tasks, and LoRA merging, it generally improves or preserves base accuracy
and yields the best model in every setting.
\end{abstract}

\section{Introduction}
\label{sec:intro}

\begin{figure}[t]
\centering
\includegraphics[width=0.95\linewidth]{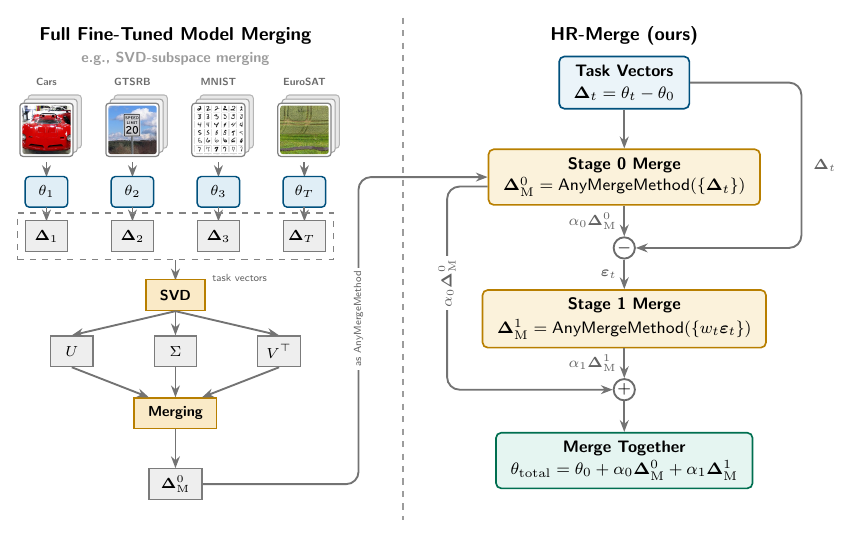}
\caption{The \proj{} pipeline. \proj{} wraps a base merging method, re-merging
headroom-weighted residuals and adding them back at a fail-closed strength.}
\label{fig:pipeline}
\end{figure}

Fine-tuning a pre-trained model for each task yields one expert per task, each
a full copy of the weights or an adapter on top of them. Storing and serving
many experts is costly, especially for large language models. Model merging
combines the experts into a single multi-task model without joint
training~\citep{wortsman2022soups,ilharco2023task,yang2024survey}. Most methods
operate on task vectors, the differences between each expert and the shared
pre-trained weights~\citep{ilharco2023task}.

Merging compresses many task vectors into one update, and their parameters
compete. Existing operators reduce this interference in different ways.
Averaging methods trim or sign-elect entries~\citep{yadav2023ties}. Subspace
methods merge along the singular directions of each task
vector~\citep{gargiulo2025tsv,marczak2025iso,zhang2026dcmerge}.
LoRA merging methods first map the adapters into a shared
space~\citep{stoica2025knots,panariello2025corespace}. All of them apply one
rule to every task. The accuracy they give up, however, is uneven. When TIES
merges six LLaMA-3 8B NLI adapters, five tasks retain at least $92\%$ of
their expert accuracy, while QNLI retains $83.6\%$
(Table~\ref{tab:lora-nli-matrix}). A rule that treats every task alike
cannot direct its correction to such a task.

Two questions follow: where the lost information is, and which tasks should
receive the correction. Section~\ref{sec:motivation} answers both. The
accuracy a merge gives up is not erased. It remains in each task's residual, its task
vector minus the merged update, and a small fraction of that
residual restores most of it (Q1). Residuals from different tasks interfere
when added together, so they must be reconciled by a merge (Q2). A
task's validation headroom, the fraction of expert validation accuracy lost
by the merge, tracks refinement gain (Q3).

Building on these observations, we propose \textbf{\proj{}}, a training-free wrapper
around any merging method (Figure~\ref{fig:pipeline}). Stage~0 runs the base
merge unchanged. Stage~1 weights each task's residual by its headroom, with a
temperature that adapts to the spread of headroom across tasks, and re-merges
the weighted residuals with the base operator. The result is added back at a
strength selected on validation. The candidates include zero, so the selected
model never falls below the base merge on validation.
Section~\ref{sec:theory} proves this for any criterion.

We evaluate \proj{} on language and vision. On LLaMA-3 8B with six NLI tasks,
it raises TIES in the Core LoRA space from $93.29$ to $94.29$ normalized
accuracy, mainly by recovering QNLI ($83.63\to90.02$). On three CLIP
backbones with 8, 14 and 20 tasks, it generally improves or preserves base
accuracy, improves DC-Merge in all nine settings, and yields
the best model in every setting.
In vision, two controls attribute the gain to the headroom
signal. Assigning the same weights to random tasks keeps about a fifth of the
gain or less on average, and spending the same validation budget on the base
coefficient adds nothing.
Per-example McNemar tests reject equal accuracy in all nine vision settings
we test.

Our main contributions are:
\begin{enumerate}
\item We show that the accuracy a merge gives up remains recoverable from
task residuals, that residuals must be merged rather than summed, and that
validation headroom identifies which tasks to refine
(Section~\ref{sec:motivation}).
\item We propose \proj{}, a training-free and method-agnostic wrapper that
re-merges headroom-weighted residuals, and prove that its fail-closed
selection cannot fall below the base merge on validation
(Sections~\ref{sec:method} and~\ref{sec:theory}).
\item We evaluate \proj{} on LLaMA-3 NLI and CLIP vision, under full
fine-tuning and LoRA, with every baseline re-run under one protocol, and
support the vision gains with per-example significance tests and controls
for the weighting and the tuning budget (Section~\ref{sec:results}). We also find
that the standard DTD evaluation directory reuses training data, and we
evaluate DTD on its official test split (Appendix~\ref{app:integrity}).
\end{enumerate}

\section{Related Work}
\label{sec:related}

\paragraph{Basic and weighted merging.}
The simplest mergers average weights~\citep{wortsman2022soups} or sum task
vectors~\citep{ilharco2023task}, arithmetic that works because models
fine-tuned from a shared initialization are (linearly)
mode-connected~\citep{garipov2018mode,frankle2020lmc} with approximately
disentangled task vectors~\citep{ortiz2023tangent}. Weighted variants scale
each model or parameter by Fisher information~\citep{matena2022fisher},
closed-form regression~\citep{jin2023regmean}, test-time-adapted
coefficients~\citep{yang2024adamerging}, or coefficients fit in output space
on a calibration set~\citep{evans2026outputspace}. Both fit their
coefficients by optimization, on unlabeled test inputs or on calibration
outputs; \proj{} fits nothing by descent and selects one strength from a grid
that includes zero, so it cannot fall below its base on validation
(Section~\ref{sec:theory}). An earlier line aligns independently trained
networks~\citep{singh2020otfusion,ainsworth2023gitrebasin} or fuses
fine-tunes into a better initialization~\citep{choshen2022fusing}; our setting shares
the checkpoint, so no alignment is needed. These mergers form the weak end
of our method-agnostic study, where \proj's refinement helps most.

\paragraph{Subspace merging.}
Sparse-subspace methods trim and sign-elect task
vectors~\citep{yadav2023ties}, drop outlying and negligible
entries~\citep{davari2024breadcrumbs}, balance parameter competition~\citep{du2024pcb}, randomly drop and rescale redundant
coordinates~\citep{yu2024dare}, or localize per-task supports with binary
masks~\citep{wang2024tallmasks,he2024localize}. Low-rank-subspace methods merge in a shared
singular basis: TSV-M orthogonalizes concatenated singular
vectors~\citep{gargiulo2025tsv}, Iso-CTS flattens the spectrum over common
and task-specific subspaces~\citep{marczak2025iso}, and DC-Merge preserves
the directional consistency of the singular
geometry~\citep{zhang2026dcmerge}; a parallel line merges low-rank (LoRA)
adaptations in a shared alignment basis or by rank-wise
clustering~\citep{stoica2025knots,panariello2025corespace,lu2026geometry,zhao2025lego}.
Optimization-based mergers minimize interference directly through data-free
objectives~\citep{cheng2025wudi,wei2025doge}. 

\paragraph{Localizing and restoring task information.}
The accuracy lost to merging is not erased but localized and recoverable:
per-task binary masks retrieve $>\!99\%$ of single-task
accuracy~\citep{wang2024tallmasks}, per-task masks with rescalers realign each expert around an elected
model~\citep{huang2024emr}, input-dependent routers recombine shared and
exclusive knowledge~\citep{lu2024twin}, and task-specific plugins correct
the merged representation~\citep{yang2024surgery}.
Figure~\ref{fig:motivation}a makes the same point one level down. The distinction is deployment: those methods keep per-task artifacts at
inference, whereas \proj{} folds a
weighted share of the residual back into \emph{one} deployable model.

\paragraph{Per-task diagnostics in the lineage.}
Lineage papers measure weights-only per-task diagnostics but never couple
one to per-task corrective action. The two that relate similarity to
performance converge on a negative result with different measurement
objects: observationally, task-vector-to-merge cosine similarity does not
track per-task normalized accuracy~\citep{marczak2025iso}; interventionally,
perturbing a task vector's singular-value energy leaves performance largely
stable while cosine similarity changes markedly~\citep{zhang2026dcmerge}.
Nor does any computed diagnostic decide \emph{which task} gets corrective
budget: directional similarity stays post hoc, the subspace alignment ratio
is motivation only, and singular-task interference is acted on globally,
never per task~\citep{zhang2026dcmerge,marczak2025iso,gargiulo2025tsv}. We
instead weight the second pass by validation headroom, a direct measurement
of the per-task deficit rather than a similarity proxy
(Section~\ref{sec:obs3}).

\section{Preliminaries}
\label{sec:preliminaries}

\paragraph{Task vectors and model merging.}
Let $\theta_0\in\mathbb{R}^{d}$ be a pre-trained model. Fine-tuning it for
task $t$ produces an expert $\theta_t$. We represent what that expert learned
by its \emph{task vector},
\begin{equation}
\Delta_t = \theta_t - \theta_0.
\label{eq:task-vector}
\end{equation}
Given $T$ task vectors, a model-merging method applies an operator
$\mathcal{M}$ to produce one shared update,
\begin{equation}
\Delta_{\mathrm{M}}^0 = \mathcal{M}(\{\Delta_t\}_{t=1}^{T}),
\qquad
\theta^0 = \theta_0 + \alpha_0\Delta_{\mathrm{M}}^0,
\label{eq:base-merge}
\end{equation}
where $\alpha_0$ is selected on held-out validation data. The result
$\theta^0$ is the \emph{Stage-0} merged model. This formulation covers both
full fine-tuning and LoRA adapters: in the latter case, $\Delta_t$ denotes the
adapter update rather than a full weight difference~\citep{ilharco2023task,hu2022lora}.

\paragraph{Validation headroom.}
Let $A_t(\theta)$ denote validation accuracy on task $t$ and
$a_t=A_t(\theta_t)>0$ its expert's validation accuracy. The gap is
\begin{equation}
g_t = \max\!\left(0,\,1-
\frac{A_t(\theta^0)}{A_t(\theta_t)}\right).
\label{eq:headroom-def}
\end{equation}
We call $g_t$ the task's \emph{validation headroom}. It is the fraction of the
expert's validation accuracy that the base merge has not retained: $g_t=0$
means that there is no measured deficit to repair, while a larger value means
that task has more room to improve.

\paragraph{Residual task information.}
For task $t$, the part of its update not represented by the Stage-0 merge is
the residual
\begin{equation}
\varepsilon_t = \Delta_t - \alpha_0\Delta_{\mathrm{M}}^0.
\label{eq:residual}
\end{equation}
The next section shows that these residuals still contain recoverable
task-specific information, but residuals from different tasks can interfere.
\proj{} therefore re-merges them rather than adding them, and uses the
headroom $g_t$ to give more weight to tasks with a larger measured deficit.

\section{Motivation}
\label{sec:motivation}

Conventional model merging compresses task-specific updates into a single
shared model. In doing so, it can leave some task information behind. A
further refinement then raises three research questions: \textbf{Q1.} \emph{Can
the missing information be recovered?} \textbf{Q2.} \emph{Can
information from different tasks be restored jointly without introducing new
interference?} \textbf{Q3.} \emph{Which tasks should the refinement
prioritize?} We address these questions through three empirical studies.

\begin{figure*}[t]
\centering
\begin{minipage}[b]{0.313\textwidth}
\centering
\includegraphics[width=\linewidth]{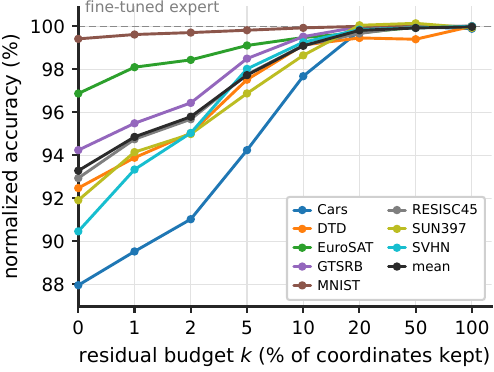}\\[10pt]
{\small\hspace{20pt}(a)}
\end{minipage}\hspace{6pt}
\begin{minipage}[b]{0.302\textwidth}
\centering
\includegraphics[width=\linewidth]{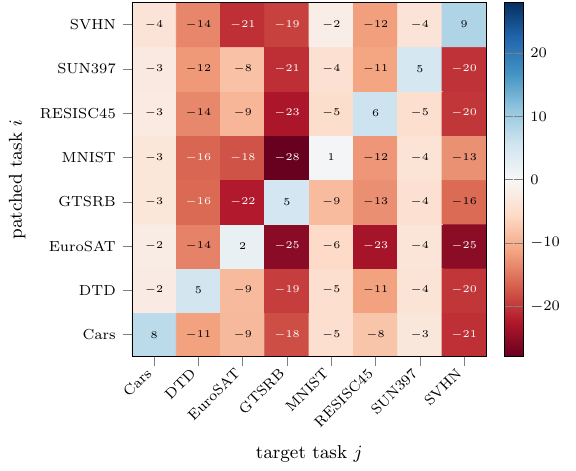}\\[1pt]
{\small\hspace{29pt}(b)}
\end{minipage}\hspace{6pt}
\begin{minipage}[b]{0.295\textwidth}
\centering
\includegraphics[width=\linewidth]{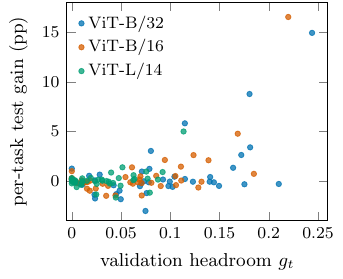}\\[5.5pt]
{\small\hspace{37pt}(c)}
\end{minipage}
\caption{The three motivating observations, all on the DC-Merge base.
(a)~Accuracy recovery versus residual budget $k$ (normalized, ViT-B/32,
8 tasks): the sacrificed accuracy is recoverable from a small fraction of
the residual. (b)~Cross-task interference: each entry $(i,j)$ is the
accuracy change on task $j$ from adding back task $i$'s top-$10\%$
residual; self-patches (diagonal, blue) help, while cross-task patches (red)
hurt. (c)~Gain under headroom weighting: one point per task, validation
headroom $g_t$ against test gain, pooled over nine cells and colored by
backbone.}
\label{fig:motivation}
\end{figure*}

\subsection{Task Information Remains Recoverable in the Residual}
\label{sec:recoverability}

\observationbox{1}{The accuracy sacrificed by a Stage-0 merge is not erased but
can be recovered from a small fraction of each task's residual.}
To examine this finding, we conduct a controlled restoration experiment for
each task. For task $t$, the residual $\varepsilon_t$ contains the component
of its expert update that the Stage-0 merge does not incorporate. We rank its
coordinates by absolute value and let $m_t^{(k)}$ denote the binary mask that
retains the top-$k\%$ coordinates. The patched model is
\begin{equation}
\hat{\theta}_t^{(k)}
= \theta_0 + \alpha_0\Delta_{\mathrm{M}}^0
+ \bigl(m_t^{(k)} \odot \varepsilon_t\bigr),
\label{eq:mot-probe}
\end{equation}
We evaluate $\hat{\theta}_t^{(k)}$ only on task $t$ and sweep $k$ over
increasing residual budgets.

Figure~\ref{fig:motivation}a reports this experiment on the ViT-B/32 eight-task
suite. The uncorrected Stage-0 merge reaches an average normalized accuracy
of $93.3\%$. Reinstating the top $10\%$ of each task's residual raises the
average to $99.1\%$, and the top $20\%$ reaches $99.8\%$. The recovery is
therefore concentrated in a small set of high-magnitude residual~coordinates of each task.

This task-wise experiment shows that residuals provide a compact source of
recoverable task information. Realizing this potential in a single shared
model requires resolving interactions among residuals from different~tasks.

\subsection{Task Residuals Interfere Across Tasks}
\label{sec:interference}
\observationbox{2}{Task-specific residuals restore their source tasks but
interfere with other tasks.}
All measurements in this observation use the ViT-B/32 eight-task suite with
DC-Merge as the Stage-0 base.
To test compatibility across task residuals, we select the top-$10\%$
coordinates by absolute value from the residual of each source task $i$ and
add the resulting update to the Stage-0 model. We then evaluate not only the
source task $i$, but every target task $j$. Each entry $(i,j)$ in
Figure~\ref{fig:motivation}b gives the accuracy change on task $j$ after
patching the residual of task $i$. Diagonal entries measure recovery on the
source task, and off-diagonal entries measure cross-task effects.
All eight diagonal entries are positive, ranging from $+0.5$ to $+8.6$
points with a mean gain of $+5.1$ points. In contrast, all 56 off-diagonal
entries are negative, with a mean change of $-11.7$ points. Patching the
MNIST residual into GTSRB produces the largest drop of $27.9$ points. These
results establish the task specificity of the information carried by each
residual. Both motivating observations replicate on ViT-B/16
(Appendix~\ref{app:b16}).
Residuals are therefore concentrated and recoverable sources of task
information, but they cannot be added directly to a shared model. The second
stage must use a merging operator to reconcile conflicts among task residuals.
Since tasks need not benefit equally from this correction, the next finding
introduces a per-task signal for allocating the second-stage update.

\subsection{Validation Headroom Identifies Refinement Priority}
\label{sec:obs3}
\observationbox{3}{Under headroom weighting, the gain concentrates on
high-headroom tasks.}
Stage-0 merging leaves different tasks with different accuracy deficits, so
refinement needs a per-task prioritization signal. The validation headroom
$g_t$ in Eq.~\ref{eq:headroom-def} measures the fraction of task $t$'s expert
accuracy that Stage-0 fails to retain. A larger $g_t$ indicates a larger
deficit. The signal reuses the validation evaluations that select Stage-0 and
requires neither additional training nor task-specific diagnosis.
Observation~2 shows that a large deficit need not be recoverable, because task
residuals can interfere. We therefore test whether headroom identifies the
tasks that benefit from refinement. Across nine DC-Merge
backbone--task-scale configurations~\citep{zhang2026dcmerge}, we first compute
each task's validation headroom on the Stage-0 model. We then weight all
residuals by their headroom scores and re-merge them jointly. For every task,
we measure the resulting test-accuracy change relative to Stage-0. Each point
in Figure~\ref{fig:motivation}c represents one task. Its horizontal coordinate
is the pre-refinement headroom, and its vertical coordinate is the realized
test gain. We compute a Spearman correlation within each configuration.
All nine correlations are positive. Their mean Spearman coefficient is
$\rho=+0.347$. Tasks that lose more ability in the initial merge tend to gain
more from joint refinement, and their larger deficits are also the more
recoverable ones. With uniform weights the association disappears
(Appendix~\ref{app:recoverability}). \proj{} therefore weights the
residuals by headroom, then re-merges them with a merging operator rather than summing
them directly, as required by Observation~2.

\section{Method}
\label{sec:method}

\subsection{Two-Stage Residual Refinement}
\label{sec:two-stage}

The three observations motivate a second refinement stage. Task information
remains in the residuals, residuals from different tasks must be reconciled,
and tasks with larger deficits merit more corrective budget. \proj{} wraps a
base merging method in a two-stage residual refinement process, as shown in
Figure~\ref{fig:pipeline} and Algorithm~\ref{alg:hr-merge}.
Stage-0 is the base merge in Eq.~\ref{eq:base-merge}. The base operator
$\mathcal{M}$ produces the shared update $\Delta_{\mathrm{M}}^0$. Validation
selects its scale $\alpha_0$ and yields the base model $\theta^0$. \proj{}
leaves this stage unchanged and requires no retraining of the task experts.

\begin{algorithm}[t]
\caption{Headroom-weighted residual refinement in \proj{}}
\label{alg:hr-merge}
\scriptsize
\setlength{\abovedisplayskip}{1pt}
\setlength{\belowdisplayskip}{1pt}
\setlength{\abovedisplayshortskip}{1pt}
\setlength{\belowdisplayshortskip}{1pt}
\setlength{\jot}{1pt}
\KwIn{task vectors $\{\Delta_t\}_{t=1}^{T}$; base operator $\mathcal{M}$;
validation grids for $\alpha_0$ and $\alpha_1$; temperature constant $c$;
expert validation accuracies $a_t$.}
\KwOut{merged model $\theta^\star$.}
\noindent$\triangleright$~\textbf{Stage-0 base merge}
\begin{align}
\Delta_{\mathrm{M}}^0 &= \mathcal{M}\!\left(\{\Delta_t\}_{t=1}^{T}\right),
\qquad
\theta^0 = \theta_0 + \alpha_0\Delta_{\mathrm{M}}^0 .
\end{align}
\noindent$\triangleright$~\textbf{Headroom-weighted residual re-merge}
\begin{align}
g_t &= \max\!\left(0,\,1-
\frac{A_t(\theta^0)}{A_t(\theta_t)}\right), \\
\varepsilon_t &= \Delta_t - \alpha_0\Delta_{\mathrm{M}}^0, \\
\tau &= c\cdot\operatorname{std}(\{g_t\}_{t=1}^{T}), \\
w_t &= T\cdot\operatorname{softmax}(\mathbf{g}/\tau)_t, \\
\Delta_{\mathrm{M}}^1 &=
\mathcal{M}\!\left(\{w_t\varepsilon_t\}_{t=1}^{T}\right).
\end{align}
\noindent$\triangleright$~\textbf{Refinement strength selection}
\begin{align}
\alpha_1^\star
&= \arg\max_{\alpha_1\in\{0\}\cup\mathcal{G}}
\frac{1}{T}\sum_{t=1}^{T} a_t^{-1}A_t\!\left(
\theta_0+\alpha_0\Delta_{\mathrm{M}}^0 \,\right.\nonumber\\[-0.2em]
&\hspace{9.8em}\left.{}+\alpha_1\Delta_{\mathrm{M}}^1\right), \\
\theta^\star &= \theta_0+\alpha_0\Delta_{\mathrm{M}}^0+
\alpha_1^\star\Delta_{\mathrm{M}}^1 .
\end{align}
\end{algorithm}

Stage-1 starts from the task residuals $\varepsilon_t$ defined in
Eq.~\ref{eq:residual}. It operates only on what Stage-0 did not retain. The next two subsections
specify the weights, the re-merge, and the refinement strength.

\subsection{Headroom-Weighted Residual Re-Merge}
\label{sec:headroom-weighting}

The headroom $g_t$ measures the fraction of task $t$'s expert validation
accuracy not retained by Stage-0 (Eq.~\ref{eq:headroom-def}). We map
headroom to Stage-1 weights via
\begin{align}
\tau &= c\cdot\operatorname{std}(\{g_t\}_{t=1}^{T}), \\
w_t &= T\cdot\operatorname{softmax}(\mathbf{g}/\tau)_t.
\label{eq:headroom-weights}
\end{align}
Here, $c$ is a temperature constant. The factor $T$ normalizes the mean
weight to one. The weighting therefore changes the allocation of the
second-stage update across tasks without changing its overall scale. A larger
$g_t$ receives a larger $w_t$ and gives task $t$ more influence in Stage-1.

The temperature adapts to the dispersion of headroom across tasks. We select
$c$ on one validation configuration for each base method, then fix it across
the remaining backbones and task scales for that method. Equal headroom scores
produce uniform weights, $w_t=1$.

We re-merge the weighted residuals with the merging operator,
\begin{equation}
\Delta_{\mathrm{M}}^1 =
\mathcal{M}\!\left(\{w_t\varepsilon_t\}_{t=1}^{T}\right).
\label{eq:refinement-merge}
\end{equation}
rather than summing them directly. This choice addresses the cross-task
interference in Observation~2. Stage-1 uses the Stage-0 operator by default.
Thus, \proj{} can wrap different model-merging methods without changing their
merging rules.

\subsection{Refinement Strength Selection}
\label{sec:strength-selection}

We select the strength $\alpha_1$ of $\Delta_{\mathrm{M}}^1$ on validation
from $\{0\}\cup\mathcal{G}$, maximizing mean accuracy normalized by
expert validation accuracy $a_t$,
\begin{align}
\alpha_1^\star
&= \arg\max_{\alpha_1\in\{0\}\cup\mathcal{G}}
\frac{1}{T}\sum_{t=1}^{T} a_t^{-1}A_t\!\left(
\theta_0 + \alpha_0\Delta_{\mathrm{M}}^0 \,\right. \nonumber\\[-0.2em]
&\hspace{9.8em}\left.{}+ \alpha_1\Delta_{\mathrm{M}}^1\right).
\label{eq:refinement-strength}
\end{align}
The final model is
\begin{equation}
\theta^\star = \theta_0 + \alpha_0\Delta_{\mathrm{M}}^0 +
\alpha_1^\star\Delta_{\mathrm{M}}^1.
\label{eq:hr-merge-output}
\end{equation}
This is a one-dimensional search. Its candidate set includes $\alpha_1=0$,
which retains the Stage-0 model if refinement receives no validation support.
The next section analyzes the resulting validation property and its scope.

\section{Theoretical Analysis}
\label{sec:theory}
\begin{table*}[t]
\centering
\small
\setlength{\tabcolsep}{11.1pt}
\begin{tabular}{llcccccc|l}
\toprule
\textbf{Method} & \textbf{LoRA space} & SNLI & MNLI & SICK & QNLI & RTE & SciTail & \textbf{Avg} \\
\midrule
\multicolumn{2}{l}{\textit{Abs.\ accuracy} (experts)} & 92.50 & 90.31 & 91.58 & 94.49 & 89.86 & 96.52 & 92.54 \\
\midrule
TSV & Full & 94.71 & 97.15 & 84.29 & 85.20 & 99.19 & 97.12 & 92.94 \\
\rowcolor{oursblue}
\textbf{+\,HR} & Full & 94.82 & 97.32 & 84.87 & 85.66 & 99.19 & 97.17 & \textbf{93.17}\,{\scriptsize(+0.23)} \\
\midrule
TIES & Full & 95.48 & 96.31 & 83.71 & 69.56 & 100.81 & 96.93 & 90.46 \\
 & KnOTS & 87.89 & 94.09 & 90.34 & 85.03 & 100.00 & 97.03 & 92.40 \\
 & Core & 92.12 & 93.54 & 93.43 & 83.63 & 99.19 & 97.81 & 93.29 \\
\rowcolor{oursblue}
\textbf{+\,HR} & Core & 92.13 & 92.91 & 94.30 & 90.02 & 99.19 & 97.17 & \textbf{94.29}\,{\scriptsize(+1.00)} \\
\bottomrule
\end{tabular}
\caption{Per-task test normalized accuracy on the six-task NLI LoRA
benchmark with LLaMA-3 8B. Shaded rows apply \proj{} to the base above them;
parentheses give the gain in average normalized accuracy.}
\label{tab:lora-nli-matrix}
\end{table*}

Every data-dependent choice in \proj{} is made on validation data from a
finite candidate set $\mathcal{C}$: the base scale $\alpha_0$, the temperature $c$,
and the refinement strength $\alpha_1\in\{0\}\cup\mathcal{G}$
(Appendices~\ref{app:candidate-set} and~\ref{app:hyperparams}).

\begin{proposition}[Fail-closed selection]
\label{prop:failclosed}
Let $\Phi$ be any criterion computed from the per-task validation accuracies
$\left(A_1,\dots,A_T\right)$, and let $\alpha_1^\star$ maximize $\Phi$ over
$\{0\}\cup\mathcal{G}$ as in Eq.~\ref{eq:refinement-strength}. Then
\begin{equation}
\Phi(\theta^\star)\;\ge\;\Phi(\theta^0).
\label{eq:failclosed}
\end{equation}
\end{proposition}

\noindent\emph{Proof.} The candidate $\alpha_1=0$ yields exactly $\theta^0$
in Eq.~\ref{eq:hr-merge-output}, and $0\in\{0\}\cup\mathcal{G}$. The maximum
over the candidate set is therefore at least the value at $\alpha_1=0$.
\hfill$\square$

\medskip
\noindent Eq.~\ref{eq:refinement-strength} uses mean validation accuracy
normalized by expert validation accuracy $a_t$.

The guarantee holds on validation. Its transfer to the test distribution
depends on how often the two models disagree on a validation example,
\begin{equation}
\hat\rho = \frac{1}{n}\sum_{i=1}^{n}
\mathbf{1}\!\left[\theta^\star(x_i)\neq\theta^0(x_i)\right],
\label{eq:disagreement}
\end{equation}
over the $n$ selection examples.

\begin{theorem}[Transfer of the guarantee]
\label{thm:transfer}
Under Assumptions~\ref{as:iid} and~\ref{as:split}
(Appendix~\ref{app:assumptions}), for any $\delta\in(0,1)$, with probability
at least $1-\delta$,
\begin{align}
A(\theta^\star) &\;\ge\; A(\theta^0)-\varepsilon,
\label{eq:transfer} \\
\varepsilon &= \sqrt{\frac{2\hat\rho L}{n-1}}+\frac{14L}{3(n-1)},
\label{eq:slack}
\end{align}
with $L=\ln\left(2|\mathcal{C}|/\delta\right)$ and $A$ the expected accuracy.
\end{theorem}

\noindent The two assumptions are that validation and test examples come from
the same distribution, and that the Stage-1 weights are computed on a split
disjoint from the one that selects $\alpha_1$. Appendix~\ref{app:transfer}
proves the theorem. Our runs
use one validation split for both roles, so Assumption~\ref{as:split}
describes a split variant of the pipeline rather than the configuration
behind Table~\ref{tab:method-agnostic}.
On our validation splits, $\hat\rho$ ranges from $0.9$ to $4.2\%$, and the
slack of the multi-task bound is $8$ to $16$ points (Appendix~\ref{app:slack}),
larger than every gain we observe. It bounds what the selection can cost on
test; it cannot certify a gain.

\section{Results}
\label{sec:results}
\begin{table*}[t]
\centering
\newcommand{\tabacc}[2]{\ensuremath{#1_{(#2)}}}
\newcommand{\tabbest}[2]{\ensuremath{\mathbf{#1}_{(\mathbf{#2})}}}
\resizebox{\textwidth}{!}{%
\begin{tabular}{llccc|ccc|ccc}
\toprule
\multicolumn{2}{l}{\textbf{Method}} & \multicolumn{3}{c}{ViT-B/32} & \multicolumn{3}{c}{ViT-B/16} & \multicolumn{3}{c}{ViT-L/14} \\
\cmidrule(lr){3-5}\cmidrule(lr){6-8}\cmidrule(lr){9-11}
 & & 8 tasks & 14 tasks & 20 tasks & 8 tasks & 14 tasks & 20 tasks & 8 tasks & 14 tasks & 20 tasks \\
\midrule
\multicolumn{2}{l}{\textit{Zero-shot}}
 & 48.12 & 57.12 & 56.04 & 55.45 & 61.32 & 59.74 & 65.39 & 68.59 & 65.50 \\
\multicolumn{2}{l}{\textit{Individual}}
 & 90.51 & 89.56 & 90.45 & 92.68 & 91.64 & 92.39 & 94.10 & 93.33 & 94.05 \\
\midrule
Weight Avg. & Base & \tabacc{65.34}{72.6} & \tabacc{64.13}{71.7} & \tabacc{60.86}{67.9} & \tabacc{71.82}{77.4} & \tabacc{69.41}{75.4} & \tabacc{65.37}{70.9} & \tabacc{79.19}{84.1} & \tabacc{76.77}{81.8} & \tabacc{71.67}{76.2} \\
 & +\,HR & \tabbest{67.02}{74.2} & \tabbest{66.89}{74.3} & \tabacc{60.86}{67.9} & \tabbest{72.53}{78.2} & \tabbest{70.94}{77.1} & \tabbest{65.60}{70.9} & \tabbest{80.52}{85.5} & \tabbest{78.13}{83.5} & \tabbest{75.70}{80.2} \\
\midrule
Task Arith. & Base & \tabacc{69.99}{77.1} & \tabacc{65.09}{72.6} & \tabacc{61.21}{68.2} & \tabacc{75.50}{81.1} & \tabacc{70.59}{76.6} & \tabacc{65.85}{71.4} & \tabacc{83.90}{88.9} & \tabacc{79.47}{84.7} & \tabacc{73.89}{78.4} \\
 & +\,HR & \tabacc{69.99}{77.1} & \tabbest{66.29}{73.9} & \tabacc{61.21}{68.2} & \tabbest{75.75}{81.6} & \tabacc{70.59}{76.6} & \tabacc{65.85}{71.4} & \tabacc{83.90}{88.9} & \tabacc{79.47}{84.7} & \tabbest{74.19}{78.7} \\
\midrule
TIES & Base & \tabacc{72.59}{80.0} & \tabacc{67.48}{75.1} & \tabacc{62.95}{70.0} & \tabacc{77.58}{83.4} & \tabacc{72.56}{78.7} & \tabacc{67.67}{73.2} & \tabacc{84.06}{89.1} & \tabacc{79.94}{85.3} & \tabacc{75.63}{80.2} \\
 & +\,HR & \tabacc{72.59}{80.0} & \tabbest{68.00}{75.6} & \tabacc{62.95}{70.0} & \tabacc{77.58}{83.4} & \tabacc{72.56}{78.7} & \tabbest{67.99}{73.4} & \tabbest{84.10}{89.3} & \tabbest{80.01}{85.6} & \tabbest{77.00}{81.6} \\
\midrule
TSV-M & Base & \tabacc{83.61}{92.0} & \tabacc{79.37}{88.3} & \tabacc{76.59}{84.5} & \tabacc{87.24}{93.9} & \tabacc{83.70}{91.0} & \tabacc{80.24}{86.7} & \tabacc{91.30}{96.9} & \tabacc{88.22}{94.3} & \tabacc{87.18}{92.5} \\
 & +\,HR & \tabacc{83.61}{92.0} & \tabbest{79.83}{88.8} & \tabbest{77.80}{85.9} & \tabacc{87.24}{93.9} & \tabbest{84.18}{91.6} & \tabbest{81.76}{88.2} & \tabbest{91.43}{97.1} & \tabbest{88.51}{94.7} & \tabbest{88.14}{93.6} \\
\midrule
Iso-CTS & Base & \tabacc{83.62}{92.3} & \tabacc{80.30}{89.4} & \tabacc{77.28}{85.3} & \tabacc{89.07}{96.0} & \tabacc{84.99}{92.4} & \tabacc{81.60}{88.2} & \tabacc{92.94}{98.7} & \tabacc{89.70}{95.9} & \tabacc{89.30}{94.7} \\
\multicolumn{2}{r}{(TSV-M) +\,HR\hphantom{$^{\dagger}$}} & \tabbest{84.11}{92.8} & \tabbest{81.15}{90.4} & \tabbest{78.94}{87.1} & \tabbest{89.36}{96.3} & \tabbest{85.60}{93.2} & \tabbest{83.10}{89.7} & \tabbest{93.10}{98.9} & \tabbest{90.47}{96.9} & \tabbest{89.89}{95.5} \\
\midrule
DC-Merge & Base & \tabacc{84.70}{93.3} & \tabacc{81.49}{90.7} & \tabacc{80.13}{88.5} & \tabacc{88.96}{95.9} & \tabacc{85.83}{93.5} & \tabacc{84.13}{91.0} & \tabacc{92.42}{98.2} & \tabacc{90.14}{96.5} & \tabacc{90.06}{95.7} \\
\rowcolor{oursblue}
 & +\,HR$^{\dagger}$ & \tabbest{85.21}{93.9} & \tabbest{82.23}{91.6} & \tabbest{80.92}{89.3} & \tabbest{89.17}{96.1} & \tabbest{86.23}{93.9} & \tabbest{85.03}{91.9} & \tabbest{92.55}{98.3} & \tabbest{90.24}{96.6} & \tabbest{90.22}{95.9} \\
\bottomrule
\end{tabular}
}
\caption{Test-split absolute accuracy under full fine-tuning. Parenthesized
subscripts report normalized accuracy. Six base methods are shown with their \proj{}
variants. Bold marks an improvement over the paired base. Identical paired cells
return the base model with
$\alpha_1^{\star}=0$. The symmetric Iso-CTS variant returns the base in every
setting and is omitted. The dagger and shading mark the main configuration. Zero-shot and
Individual are our evaluations.}
\label{tab:method-agnostic}
\end{table*}

\subsection{Experimental Setup}
\label{sec:experimental-setup}

\paragraph{Models and backbones.}
We evaluate model merging under full fine-tuning and LoRA adaptation. For
full fine-tuning, we use the released experts of three CLIP encoders,
ViT-B/32, ViT-B/16, and
ViT-L/14~\citep{radford2021clip,wang2024tallmasks}. The LoRA study uses
released rank-16 adapters for LLaMA-3 8B and CLIP
ViT-B/32~\citep{llama3herd,stoica2025knots}.

\paragraph{Datasets.}
Our language study uses the six-task NLI benchmark of
\citet{stoica2025knots} on LLaMA-3 8B. It contains SNLI, MNLI, SICK, QNLI,
RTE, and SciTail. The vision study follows public task-vector image
classification benchmarks~\citep{ilharco2023task,wang2024tallmasks}. The
full fine-tuning setting covers nested suites of 8, 14, and 20 tasks. The LoRA
setting uses the eight-task suite on ViT-B/32. We retain the fixed validation
and test splits and preprocessing. Headroom and selection use expert
validation accuracies; test scores use expert test accuracies. We average
normalized scores across tasks.

\paragraph{Baselines and controlled reproduction.}
The full fine-tuning study covers six base operators. They include weight
averaging~\citep{wortsman2022soups}, Task
Arithmetic~\citep{ilharco2023task}, TIES~\citep{yadav2023ties},
TSV-M~\citep{gargiulo2025tsv}, Iso-CTS~\citep{marczak2025iso}, and
DC-Merge~\citep{zhang2026dcmerge}. The language LoRA study evaluates TSV in
the Full LoRA space and TIES in the Full, KnOTS, and Core LoRA spaces. The vision LoRA
study evaluates TSV+Iso-C and CART+Iso-C~\citep{choi2024cart} in the same
three spaces~\citep{stoica2025knots,panariello2025corespace}. All
base--wrapper comparisons come from paired runs in this shared pipeline.
We evaluate DTD on its official \texttt{test1} split. Appendix~\ref{app:integrity}
documents the inherited split issue.

\paragraph{Hyperparameter selection.}
Each base operator retains its published procedure for selecting
$\alpha_0$. We select the refinement strength $\alpha_1$ on validation data
only; on ViT-B/32 with 8 to 20 tasks, half of it retains the gain
(Appendix~\ref{app:ablation-details}). Appendix~\ref{app:setup} reports the
task compositions, protocols, coefficients, and search spaces;
Appendices~\ref{app:budget} and~\ref{app:complexity} give the runtime
environments, budget, and complexity.

\subsection{Main Results}

\subsubsection{Results on NLP Tasks}
\label{sec:exp-language}

Table~\ref{tab:lora-nli-matrix} reports base TSV in the Full LoRA space and
base TIES in all three LoRA merging spaces. Applying \proj{} to TSV raises the average
normalized accuracy from $92.94$ to $93.17$, a gain of $+0.23$ points. Five
of the six tasks improve, most on SICK ($+0.58$) and QNLI ($+0.46$), and RTE
remains unchanged.

For TIES, Core space gives the strongest base result at $93.29$. \proj{}
raises it to $94.29$, yielding a $+1.00$-point improvement and the best result
in the table. The gain is concentrated on QNLI, whose normalized accuracy
rises from $83.63$ to $90.02$. Changes on the other five tasks remain within
$0.87$ points. Appendix~\ref{app:nli-pertask} gives the headroom scores and
weights, and Appendix~\ref{app:hyperparams} the remaining cells: the strongest base,
TSV in Core space ($94.19$), is returned unchanged and lies $0.10$ below the best
model.

\subsubsection{Results on Vision Tasks}
\label{sec:exp-vision}

\paragraph{Full fine-tuning.}
\label{sec:vision-full-finetuning}

Table~\ref{tab:method-agnostic} compares six base mergers with their
\proj{} variants. The reported test accuracy improves or remains unchanged
in each cell. TSV-M on ViT-B/16 with eight tasks returns the base model
($\alpha_1^{\star}=0$). DC-Merge improves in all nine backbone--task-scale
settings. Across the six unsaturated ViT-B/32 and ViT-B/16 settings, the
mean gain is $+0.64$ normalized points. All nine changes are positive.
On the test images where \proj{} and its DC-Merge base disagree, \proj{} is
right $64$--$73\%$ of the time, and an exact McNemar test rejects equal
accuracy in all nine settings we test ($p<10^{-16}$), as does a
task-stratified test of the mean. The gains concentrate
on a few tasks, SVHN above all. Without SVHN, four settings stay
significant, the three eight-task settings do not, and the 20-task settings
on ViT-B/32 and ViT-L/14 reverse (Appendix~\ref{app:significance}). The worst
task rises in all twelve settings (Appendix~\ref{app:imbalance}).

The gains grow with the task count on ViT-B/32 and ViT-B/16. The DC-Merge
variant ranks first in six settings. The asymmetric Iso-CTS variant uses
TSV-M in Stage~1 and ranks first in the other three. Thus, the best model in
every setting is produced by \proj{}. Gains on ViT-L/14 are smaller and
range from $+0.11$ to $+0.17$. Its base models already approach the task expert ceiling: with eight tasks,
\proj{} comes within one point of a jointly trained multi-task model
\citep[93.5;][]{yang2024adamerging}, as Appendix~\ref{app:mtl-reference} details.

The pattern extends beyond the strongest base. Weight averaging gains
$+1.54$ normalized points on average, the largest improvement among the six
methods. TIES activates refinement in five of nine settings and returns the
base model in the other four. Symmetric Iso-CTS returns the base in all nine settings and is omitted from
Table~\ref{tab:method-agnostic}. Its TSV-M Stage~1 variant improves all nine. Appendix~\ref{app:failure}
details these selective cases. Appendix~\ref{app:pertask-20t} reports the
20-task per-task results.

\paragraph{LoRA merging.}
\label{sec:vision-lora}

\begin{figure}[t]
\centering
\includegraphics[width=\columnwidth]{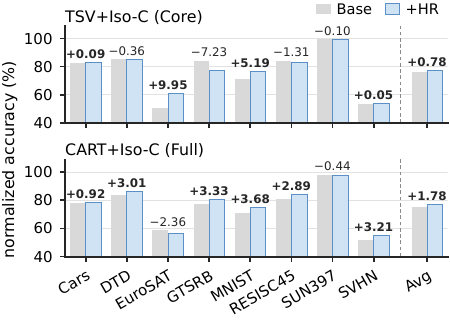}
\caption{Per-task test normalized accuracy of two vision LoRA cells on
ViT-B/32, with the change above each pair (bold for gains).
Table~\ref{tab:lora-vision-all} lists all twelve cells.}
\label{fig:lora-vision}
\end{figure}

Figure~\ref{fig:lora-vision} shows the two headline Vision LoRA results.
\proj{} raises TSV+Iso-C in Core space from $76.42$ to $77.20$, a gain of
$+0.78$ normalized points. EuroSAT and MNIST gain $+9.95$ and $+5.19$
points. GTSRB declines by $7.23$ points, the largest loss.

CART+Iso-C obtains a larger gain in Full space, from $74.94$ to $76.73$
($+1.78$ points), and six of the eight tasks improve. The KnOTS and Core cells select $\alpha_1^{\star}=0$ and
return their base models (coefficients in
Appendix~\ref{app:hyperparams}).

\subsection{Ablations and Analysis}
\label{sec:ablations-analysis}

\paragraph{Stage-1 weighting.}
Table~\ref{tab:ladder} isolates the Stage-1 weights on ViT-B/16. All rows
share the base merge, the residual re-merge and the strength selection, and
differ only in $w_t$. Uniform weights treat every task alike. Hard selection
refines only tasks below the average validation accuracy.
The two soft rows apply the headroom softmax with a fixed temperature
$\tau=0.2$ or with the adaptive $\tau=c\cdot\operatorname{std}(g)$ of
\proj{}. The adaptive temperature gives the largest gain in every column.
The one inversion is the fixed temperature, which falls below hard selection
at 20 tasks. A temperature tuned once does not scale with the number of
tasks.

\begin{table}[t]
\centering
\footnotesize
\setlength{\tabcolsep}{5pt}
\begin{tabular}{lccc}
\toprule
 & \multicolumn{3}{c}{ViT-B/16} \\
\cmidrule(lr){2-4}
Stage-1 weighting & 8 tasks & 14 tasks & 20 tasks \\
\midrule
uniform ($w_t{=}1$)              & $+0.03$ & $+0.13$ & $+0.01$ \\
hard selection                   & $+0.09$ & $+0.18$ & $+0.33$ \\
soft, fixed $\tau{=}0.2$         & $+0.14$ & $+0.22$ & $+0.31$ \\
headroom, permuted$^{\dagger}$   & $+0.04$ & $+0.10$ & $+0.02$ \\
\rowcolor{oursblue}
\proj{} (adaptive $\tau$)        & $\mathbf{+0.23}$ & $\mathbf{+0.45}$ & $\mathbf{+0.92}$ \\
\bottomrule
\end{tabular}
\caption{Stage-1 weighting ablation on ViT-B/16: gain in normalized accuracy
over the DC-Merge base. $^{\dagger}$Mean over three random reassignments of the
\proj{} weights across tasks (Appendix~\ref{app:ablation-details}).}
\label{tab:ladder}
\end{table}

\paragraph{Which tasks receive the weight.}
The permuted row keeps the weight values and the temperature of \proj{} but
assigns the weights to random tasks, so the task with the largest headroom no
longer receives the largest weight. Averaged over three permutations, it
gains $+0.04$ at 8 tasks against $+0.23$ for \proj{}, $+0.10$ against
$+0.45$ at 14 tasks, and $+0.02$ against $+0.92$ at 20 tasks. In five of the
nine runs, validation selects $\alpha_1^{\star}=0$ and returns the base.
The gain therefore comes from which tasks receive the residual, not from
weighting the tasks unequally.

\paragraph{A second tuned coefficient.}
Could the gain come from spending $21$ validation
evaluations on any second coefficient? With Stage~1 disabled, we spend them
on $\alpha_0\in[1.40,2.40]$ around the deployed $1.9$ for both eight-task
suites. On ViT-B/32, validation peaks at the deployed value, so the retuned
base gains $+0.00$ on test, against $+0.59$ for \proj{}. On ViT-B/16,
validation prefers $1.95$, which loses $0.09$ on test, against $+0.23$ for \proj{}.

\section{Conclusion}
\label{sec:conclusion}

We introduced \proj{}, a training-free wrapper that refines any merging
method where its merge falls short. The accuracy a merge gives up remains in
the task residuals, and validation headroom identifies which tasks to refine.
\proj{} re-merges the headroom-weighted residuals with the base operator and
selects their strength on validation, so it never falls below the base merge
on the validation objective. It yields the strongest LLaMA-3 NLI merge we
test, $94.29$, and the best model in every full fine-tuning setting,
and controls on vision trace the gain to the headroom assignment rather than
to extra tuning. Its gains are largest where the base merge leaves the most
headroom.

\section*{Limitations}

\paragraph{Scope of the language study.}
Our language experiments use the six-task NLI benchmark on LLaMA-3 8B with the
LoRA adapters released by \citet{stoica2025knots}, the setting also used by
\citet{panariello2025corespace}. Other model families and generative tasks are
left to future work.

\paragraph{Task metrics.}
Like the merging methods we compare with, we evaluate only tasks scored by
accuracy. The two-stage refinement itself does not depend on the metric:
Stage~1 re-merges weighted residuals for any choice of weights. Headroom and
the validation score that selects the strength, however, compare tasks
through their accuracy relative to the expert. Tasks scored by error metrics,
where lower is better, or by different generation metrics would need these
quantities on a common, signed scale, which we leave to future work.

\paragraph{Selection criterion.}
Like its base methods, \proj{} selects for mean accuracy, so the average can
rise while individual tasks lose accuracy (Appendix~\ref{app:lora-vision}).
Proposition~\ref{prop:failclosed} holds for any criterion computed from the
per-task validation accuracies, so a worst-task criterion can replace the mean
without changing the fail-closed guarantee. This guarantee applies to the
validation criterion and does not guarantee non-decreasing test accuracy.

\paragraph{Theory.}
The transfer bound of Theorem~\ref{thm:transfer} is stated for a variant that
computes the Stage-1 weights and selects the strength on disjoint validation
splits, as Section~\ref{sec:theory} discusses.

\paragraph{Baselines.}
LiNeS, a post-training layer rescaling evaluated on the same vision
suites~\citep{wang2025lines}, is not among our baselines, nor are
adapter-composition methods; Appendix~\ref{app:related} discusses both.

\bibliography{custom}

\appendix
\setlength{\textfloatsep}{16pt plus 2pt minus 2pt}\setlength{\dbltextfloatsep}{16pt plus 2pt minus 2pt}
\setlength{\floatsep}{10pt plus 2pt minus 2pt}\setlength{\dblfloatsep}{10pt plus 2pt minus 2pt}\setlength{\intextsep}{10pt plus 2pt minus 2pt}

\section{Replication on ViT-B/16}
\label{app:b16}

Both motivating observations of Section~\ref{sec:motivation} replicate on
ViT-B/16 (8 tasks, DC-Merge base), so neither is specific to ViT-B/32.

\begin{figure*}[t]
\begin{minipage}[t]{\columnwidth}
\centering
\includegraphics[width=0.95\columnwidth]{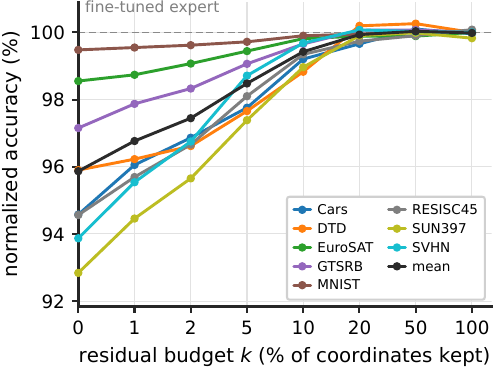}
\caption{Recovery versus residual budget $k$ on ViT-B/16 (DC-Merge base),
drawn as Figure~\ref{fig:motivation}a, with one line per task and the
mean in black.}
\label{fig:probe-b16}
\end{minipage}\hfill
\begin{minipage}[t]{\columnwidth}
\centering
\begin{tikzpicture}
\begin{axis}[
  scale only axis, width=0.48\columnwidth, height=0.48\columnwidth,
  enlargelimits=false, axis on top,
  tick align=outside, xtick pos=bottom, ytick pos=left,
  xtick={0,...,7}, ytick={0,...,7},
  xticklabels={Cars,DTD,EuroSAT,GTSRB,MNIST,RESISC45,SUN397,SVHN},
  yticklabels={Cars,DTD,EuroSAT,GTSRB,MNIST,RESISC45,SUN397,SVHN},
  x tick label style={rotate=45, anchor=east, font=\scriptsize},
  y tick label style={font=\scriptsize},
  xlabel={target task $j$}, ylabel={patched task $i$},
  label style={font=\small},
  colorbar, colorbar style={width=6pt, tick label style={font=\scriptsize}},
  colormap={RdBu}{rgb255=(103,0,31) rgb255=(178,24,43) rgb255=(214,96,77)
    rgb255=(244,165,130) rgb255=(253,219,199) rgb255=(247,247,247)
    rgb255=(209,229,240) rgb255=(146,197,222) rgb255=(67,147,195)
    rgb255=(33,102,172) rgb255=(5,48,97)},
  point meta min=-27, point meta max=27,
]
\addplot[matrix plot*, mesh/cols=8, point meta=explicit,
  nodes near coords, every node near coord/.append style={
    anchor=center, font=\tiny,
    /pgf/number format/fixed, /pgf/number format/precision=0,
    color/.expanded=\ifdim\pgfplotspointmetatransformed pt<215pt white\else black\fi,
  }]
  table[x=x, y=y, meta=d] {figs_appendix/interference_b16.dat};
\end{axis}
\end{tikzpicture}
\caption{Accuracy change (pp) on task $j$ from patching task $i$'s
top-$10\%$ residual, ViT-B/16 (DC-Merge base). Blue diagonal = self-recovery;
red off-diagonal = cross-task interference. Replicates
Figure~\ref{fig:motivation}b.}
\label{fig:interference-b16}
\end{minipage}
\end{figure*}

\paragraph{Observation 1 (concentration).}
Figure~\ref{fig:probe-b16} repeats the residual-budget sweep of
Figure~\ref{fig:motivation}a. Restoring the top $5/10/20\%$ of each task's
residual reaches a mean normalized accuracy of $98.5$/$99.4$/$99.9$,
slightly above the ViT-B/32 levels ($97.7$/$99.1$/$99.8$): the sacrificed
accuracy is at least as concentrated on the larger backbone.

\paragraph{Observation 2 (interference).}
Figure~\ref{fig:interference-b16} repeats the residual-patching matrix of
Figure~\ref{fig:motivation}b. On ViT-B/16, too, every self-patch helps
(diagonal mean $+3.3$ points) and every one of the $56$ cross-task patches
hurts (mean $-9.4$, range $-26.3$ to $-1.1$).

\section{Experimental Details}
\label{app:setup}

\subsection{Datasets and Evaluation Protocol}

The language study uses the six-task NLI benchmark released by
\citet{stoica2025knots}. It contains SNLI, MNLI, SICK, QNLI, RTE, and
SciTail. We use its rank-16 LLaMA-3 8B adapters and follow its evaluation
protocol. Each task keeps its own classification head: the heads are not
merged, and each is loaded before its task is evaluated, for every method we
compare.

The full fine-tuning vision study follows the nested task suites of
\citet{wang2024tallmasks}. The eight-task suite contains MNIST, Cars, DTD,
EuroSAT, GTSRB, RESISC45, SUN397, and SVHN. The fourteen-task suite adds
CIFAR100, STL10, Flowers102, OxfordIIITPet, FER2013, and PCAM. The twenty-task
suite further adds FashionMNIST, CIFAR10, Food101, RenderedSST2, KMNIST, and
EMNIST. The vision LoRA benchmark uses the same eight tasks with the released
rank-16 ViT-B/32 adapters of \citet{stoica2025knots}. We preserve each
benchmark's fixed validation and test splits and preprocessing. DTD is
evaluated on its official \texttt{test1} split throughout; its validation
split, which supplies the headroom and the strength selection, is the
benchmark's 376-image held-out portion of the training partition and shares
no image with \texttt{test1}.

For headroom and validation selection, the denominator is the expert
validation accuracy $a_t=A_t(\theta_t)$; the selection score is
$100\,A_t(\theta)/a_t$. Test reporting uses $100\,A_t^{\mathrm{test}}(\theta)/
A_t^{\mathrm{test}}(\theta_t)$, with expert accuracies measured on the test
split. These test denominators do not enter headroom or parameter selection.
We report the arithmetic mean over tasks. Base and wrapped models always share the same experts,
splits, preprocessing, and evaluation code.

\subsection{Full Hyperparameter Tables}
\label{app:hyperparams}

Table~\ref{tab:hyperparams-full} lists, for every base method and cell of the
full fine-tuning vision study (Section~\ref{sec:results}), the base scale
$\alpha_0$, the selected refinement strength $\alpha_1^{\star}$, and the
temperature constant $c$. The base scale follows each method's own validation
protocol. We calibrate $c$ once per base method on ViT-B/32 with 8 tasks and
then freeze it, as described in Section~\ref{sec:headroom-weighting}.
The resulting constants are $c^\star=1.4$ for DC-Merge, Task Arithmetic,
TSV-M, and both Iso-CTS variants; $c^\star=0.3$ for TIES, calibrated on
ViT-B/32 with 14 tasks through the flat-cell fallback; and $c^\star=2.4$ for
weight averaging.

The calibration sweep covers $c\in\{0.1,\allowbreak 0.2,\allowbreak\ldots,\allowbreak 3.0\}$ against a coarse
strength grid $\alpha_1\in\{0,\allowbreak 0.1,\allowbreak\ldots,\allowbreak 1.0\}$. A cell is treated as flat
if no value of $c$ clears a $0.05$-point validation margin over
$\alpha_1=0$. Deployment then uses the finer grid
$\alpha_1\in\{0,\allowbreak 0.05,\allowbreak\ldots,\allowbreak 1.0\}$ from
Section~\ref{sec:strength-selection}. The base scale $\alpha_0$ is not chosen
by \proj{}; every base method retains its published selection protocol, over
$\{1/30,2/30,\ldots,1\}$ for task arithmetic and TIES and over
$\{0.1,\ldots,3.0\}$ for the other bases. TIES runs without trimming
($k=100$), as in the code base we inherit. At deployment, a non-zero
$\alpha_1$ is kept only if it clears $\alpha_1=0$ by the same $0.05$-point
validation margin; this changes one cell, task arithmetic on ViT-B/32 with
20 tasks.

Several choices were revised during development, always on validation data:
the flat-cell fallback for TIES; the per-cell temperature and the extended
strength grid of the vision LoRA suite (Table~\ref{tab:vision-lora-hyperparams});
the TSV-M Stage~1 of Iso-CTS, introduced after the symmetric variant returned
the base in every cell (Appendix~\ref{app:failure}); and, in the language
study, the $\alpha_0$ grid, extended beyond the cap of the base matrix, and the
$\alpha_1$ grid, refined after an earlier run of the Core-space TIES cell.

\begin{table*}[t]
\centering
\footnotesize
\setlength{\tabcolsep}{4pt}
\begin{tabularx}{\textwidth}{@{}lc|YYY|YYY|YYY@{}}
\toprule
 & & \multicolumn{3}{c|}{ViT-B/32} & \multicolumn{3}{c|}{ViT-B/16} & \multicolumn{3}{c}{ViT-L/14} \\
\textbf{Base} & $c$ & 8t & 14t & 20t & 8t & 14t & 20t & 8t & 14t & 20t \\
\midrule
Weight Avg.        & 2.4 & $\tfrac{1}{T}$/.10 & $\tfrac{1}{T}$/.20 & $\tfrac{1}{T}$/.00 & $\tfrac{1}{T}$/.15 & $\tfrac{1}{T}$/.30 & $\tfrac{1}{T}$/.15 & $\tfrac{1}{T}$/.20 & $\tfrac{1}{T}$/.25 & $\tfrac{1}{T}$/.15 \\
Task Arith.        & 1.4 & .23/.00 & .13/.05 & .07/.00 & .27/.05 & .13/.00 & .07/.00 & .30/.00 & .17/.00 & .13/.05 \\
TIES               & 0.3 & 1.0/.00 & 1.0/.10 & .63/.00 & 1.0/.00 & 1.0/.00 & .93/.15 & 1.0/.10 & .97/.15 & 1.0/.10 \\
TSV-M              & 1.4 & 1.0/.00 & 0.9/.10 & 0.9/.15 & 1.0/.00 & 1.0/.10 & 1.0/.10 & 1.1/.15 & 1.0/.20 & 1.0/.20 \\
Iso-CTS (sym.\ $\mathcal{M}_1$)  & 1.4 & 1.5/.00 & 1.2/.00 & 1.1/.00 & 1.6/.00 & 1.2/.00 & 1.1/.00 & 1.9/.00 & 1.5/.00 & 1.2/.00 \\
Iso-CTS ($\mathcal{M}_1$=TSV-M)  & 1.4 & 1.5/.10 & 1.2/.15 & 1.1/.15 & 1.6/.15 & 1.2/.15 & 1.1/.10 & 1.9/.20 & 1.5/.25 & 1.2/.15 \\
DC-Merge           & 1.4 & 1.9/.55 & 1.8/.40 & 1.7/.50 & 1.9/.25 & 1.8/.35 & 1.8/.45 & 2.1/.45 & 2.2/.25 & 1.9/.55 \\
\bottomrule
\end{tabularx}
\caption{Per-cell coefficients for every base method as base scale over
refinement strength ($\alpha_0/\alpha_1^{\star}$), with the per-method
temperature constant $c$. Weight averaging fixes $\alpha_0=1/T$ by definition;
$\alpha_1^{\star}=.00$ entries are fail-closed fallbacks. TIES's $c$ is
calibrated on B/32 14-task via the flat-cell fallback rule
(Section~\ref{sec:headroom-weighting}).}
\label{tab:hyperparams-full}
\end{table*}

Table~\ref{tab:nli-hyperparams} gives the coefficients and test results of
the language cells. Its temperature constant $c=1.4$ is inherited from the
vision calibration without re-tuning on language data. \proj{} selects a
non-zero strength in seven of the ten cells and returns the base in the other
three. The strongest base is TSV in Core space ($94.19$), which \proj{}
returns unchanged; the best \proj{} model, TIES in Core space, exceeds it by
$0.10$. In this study the re-merged residual is isotropized before it is
added, as in Iso-C, although the base merge is not.

\begin{table}[t]
\centering
\small
\setlength{\tabcolsep}{2.4pt}
\begin{tabular}{llcc ccc}
\toprule
Operator & Space & $\alpha_0$ & $\alpha_1^{\star}$ & Base & +HR & $\Delta$ \\
\midrule
TSV      & Full  & 0.7 & 0.40 & 92.94 & 93.17 & $+0.23$ \\
         & KnOTS & 0.3 & 0.03 & 91.45 & 91.48 & $+0.03$ \\
         & Core  & 0.5 & 0.00 & \underline{94.19} & 94.19 & $0.00$ \\
\midrule
TIES     & Full  & 0.9 & 0.50 & 90.46 & 90.96 & $+0.50$ \\
         & KnOTS & 0.9 & 0.15 & 92.40 & 92.38 & $-0.02$ \\
         & Core  & 1.1 & 0.40 & 93.29 & \textbf{94.29} & $+1.00$ \\
\midrule
DARE-TIES & Full & 0.3 & 0.05 & 90.25 & 90.19 & $-0.06$ \\
\midrule
DC-Merge & Full  & 1.5 & 0.02 & 89.71 & 89.73 & $+0.03$ \\
         & KnOTS & 1.1 & 0.00 & 78.88 & 78.88 & $0.00$ \\
         & Core  & 1.3 & 0.00 & 86.29 & 86.29 & $0.00$ \\
\bottomrule
\end{tabular}
\caption{Coefficients and test normalized accuracy of the LLaMA-3 NLI cells.
Bold marks the best model and underline the strongest base; where validation
selects $\alpha_1^\star=0$, \proj{} returns the base. Full denotes the
complete LoRA parameter space, not full fine-tuning. We use $c=1.4$ throughout,
so $\tau=c\cdot\operatorname{std}(\mathbf{g})$ varies only with the headroom
spread of each cell.}
\label{tab:nli-hyperparams}
\end{table}

Table~\ref{tab:vision-lora-hyperparams} reports the corresponding coefficients
for the vision LoRA suite.

\begin{table}[t]
\centering
\small
\setlength{\tabcolsep}{2.4pt}
\begin{tabular}{llcccc}
\toprule
Operator & Space & $\alpha_0$ & $c$ & $\alpha_1^{\star}$ & $\Delta$ \\
\midrule
TSV+Iso-C & Full   & 4.4 & 0.1 & 0.10 & $+0.12$ \\
          & KnOTS  & 4.4 & 0.1 & 0.00 & $0.00$ \\
          & Core   & 0.9 & 0.1 & 0.40 & $+0.78$ \\
\midrule
CART+Iso-C & Full  & 3.0 & 1.9 & 0.40 & $+1.78$ \\
           & KnOTS & 2.0 & 1.9 & 0.00 & $0.00$ \\
           & Core  & 5.1 & 0.1 & 0.00 & $0.00$ \\
\midrule
TIES+Iso-C & Full  & 3.1 & 1.9 & 1.50 & $+0.87$ \\
           & KnOTS & 6.5 & 1.9 & 0.00 & $0.00$ \\
           & Core  & 2.1 & 1.9 & 0.00 & $0.00$ \\
\midrule
DC-Merge & Full  & 0.9 & 1.9 & 0.00 & $0.00$ \\
         & KnOTS & 0.1 & 1.9 & 0.10 & $+0.88$ \\
         & Core  & 0.9 & 1.9 & 0.00 & $0.00$ \\
\bottomrule
\end{tabular}
\caption{Coefficients of all twelve cells of the 8-task vision LoRA suite
(Section~\ref{sec:vision-lora}); $\Delta$ is the test normalized gain over
the paired base at the same $\alpha_0$.}
\label{tab:vision-lora-hyperparams}
\end{table}

Five cells select a nonzero refinement strength and seven return the base
model. Retesting every closed cell with $c\in\{0.1,0.7,1.9\}$, and the two
Core-space cells over $c\in\{0.1,0.3,0.6,1.0,1.4,1.9,2.5\}$, still
selects $\alpha_1^{\star}=0$ in every case.

The TIES+Iso-C Full cell reaches the boundary of the initial strength grid.
We extend the grid and evaluate $\alpha_1\in\{1.8,2.2,2.6\}$ as sentinels.
These evaluations confirm $\alpha_1^{\star}=1.5$ as an interior optimum.
CART+Iso-C merges in Full and KnOTS space use fp64.

\section{Per-Task Mechanism on NLI}
\label{app:nli-pertask}

Table~\ref{tab:nli-pertask} expands the Core-space TIES result from
Table~\ref{tab:lora-nli-matrix}. QNLI has a validation headroom of $0.160$,
more than twice that of any other task. It receives the largest residual
weight, $w_t=3.07$, and gains $+6.39$ normalized test points. Changes on the
remaining tasks have an absolute magnitude of at most $0.87$ points.

\begin{table}[t]
\centering
\small
\setlength{\tabcolsep}{4pt}
\begin{tabular}{lccccc}
\toprule
Task & Base & +\proj{} & $\Delta$ & $g_t$ & $w_t$ \\
\midrule
SNLI    & 92.12 & 92.13 & $+0.01$ & 0.078 & 0.84 \\
MNLI    & 93.54 & 92.91 & $-0.63$ & 0.061 & 0.64 \\
SICK    & 93.43 & 94.30 & $+0.87$ & 0.069 & 0.72 \\
QNLI    & 83.63 & \textbf{90.02} & $\mathbf{+6.39}$ & \textbf{0.160} & \textbf{3.07} \\
RTE     & 99.19 & 99.19 & $+0.00$ & 0.031 & 0.39 \\
SciTail & 97.81 & 97.17 & $-0.64$ & 0.022 & 0.34 \\
\midrule
\textbf{Avg} & 93.29 & \textbf{94.29} & $\mathbf{+1.00}$ & & \\
\bottomrule
\end{tabular}
\caption{Per-task breakdown of Core-space TIES on the LLaMA-3 8B NLI
benchmark. The table combines test normalized accuracy with validation
headroom $g_t$ and residual weight $w_t$.}
\label{tab:nli-pertask}
\end{table}

\section{Per-Task Results on Vision LoRA}
\label{app:lora-vision}

Table~\ref{tab:lora-vision-all} expands Figure~\ref{fig:lora-vision} to all
twelve method--space cells, adding TIES+Iso-C and DC-Merge. Their active cells
gain $+0.87$ points in Full space and $+0.88$ points in KnOTS space.

Both gains are concentrated on a few tasks. TIES+Iso-C gains $15.8$ points
on GTSRB and $10.7$ on SVHN, while MNIST and EuroSAT decline by $7.0$ and
$6.3$ points. DC-Merge gains $11.7$ points on SVHN and $7.1$ on GTSRB,
while EuroSAT declines by $7.7$ points. These changes expose the task-level
trade-offs hidden by the pool average.

\begin{table*}[t]
\centering
\resizebox{\textwidth}{!}{%
\begin{tabular}{llccccccccc}
\toprule
\textbf{Method} & \textbf{Space} & Cars & DTD & EuroSAT & GTSRB & MNIST & RESISC45 & SUN397 & SVHN & \textbf{Avg} \\
\midrule
\multicolumn{2}{l}{\textit{Abs. accuracy} (experts)} & 74.0 & 58.3 & 99.0 & 92.7 & 99.3 & 88.4 & 64.5 & 96.2 & 84.05 \\
\midrule
TSV+Iso-C & Full & 80.20 & 82.11 & 61.99 & 65.80 & 66.43 & 80.24 & 99.30 & 42.57 & 72.33 \\
 & +\,HR (Full) & 80.25 & 81.29 & 61.84 & 66.91 & 66.84 & 79.83 & 99.25 & 43.38 & \textbf{72.45}\,{\scriptsize(+0.12)} \\
 & KnOTS & 80.54 & 83.39 & 56.19 & 75.18 & 68.10 & 79.87 & 98.44 & 50.24 & 73.99 \\
 & Core & 82.93 & 85.12 & 50.88 & 84.30 & 71.03 & 84.21 & 99.29 & 53.57 & 76.42 \\
 & +\,HR (Core) & 83.02 & 84.76 & 60.83 & 77.07 & 76.22 & 82.90 & 99.19 & 53.62 & \textbf{77.20}\,{\scriptsize(+0.78)} \\
\midrule
CART+Iso-C & Full & 77.81 & 83.57 & 58.55 & 77.34 & 71.22 & 81.18 & 98.10 & 51.79 & 74.94 \\
 & +\,HR (Full) & 78.73 & 86.58 & 56.19 & 80.67 & 74.90 & 84.07 & 97.66 & 55.00 & \textbf{76.73}\,{\scriptsize(+1.78)} \\
 & KnOTS & 80.48 & 84.21 & 51.07 & 71.37 & 61.15 & 79.96 & 98.95 & 47.65 & 71.85 \\
 & Core & 83.37 & 84.39 & 50.21 & 81.85 & 71.34 & 83.05 & 99.40 & 54.25 & 75.98 \\
\midrule
TIES+Iso-C & Full & 81.95 & 75.09 & 58.29 & 38.56 & 62.84 & 69.92 & 98.98 & 40.03 & 65.71 \\
 & +\,HR (Full) & 78.42 & 74.81 & 52.00 & 54.35 & 55.80 & 69.31 & 97.22 & 50.72 & \textbf{66.58}\,{\scriptsize(+0.87)} \\
 & KnOTS & 80.22 & 80.56 & 53.16 & 66.76 & 68.71 & 77.64 & 98.34 & 50.77 & 72.02 \\
 & Core & 83.37 & 84.58 & 53.12 & 78.79 & 69.66 & 82.24 & 99.55 & 52.15 & 75.43 \\
\midrule
DC-Merge & Full & 90.83 & 79.38 & 66.74 & 54.67 & 92.22 & 75.25 & 99.12 & 64.78 & 77.87 \\
 & KnOTS & 81.99 & 75.09 & 50.51 & 41.45 & 53.94 & 71.52 & 98.40 & 40.11 & 64.12 \\
 & +\,HR (KnOTS) & 82.58 & 73.90 & 42.80 & 48.52 & 52.29 & 71.27 & 96.83 & 51.85 & \textbf{65.01}\,{\scriptsize(+0.88)} \\
 & Core & 90.73 & 79.38 & 66.70 & 54.79 & 92.06 & 75.33 & 99.13 & 64.77 & 77.86 \\
\bottomrule
\end{tabular}%
}
\caption{Per-task test normalized accuracy for all twelve cells of the
eight-task Vision LoRA suite. A \proj{} row appears for each active cell.
Cells without one select $\alpha_1^{\star}=0$ and return their base model.}
\label{tab:lora-vision-all}
\end{table*}

\begin{figure*}[t]
\centering
\includegraphics[width=0.9\textwidth]{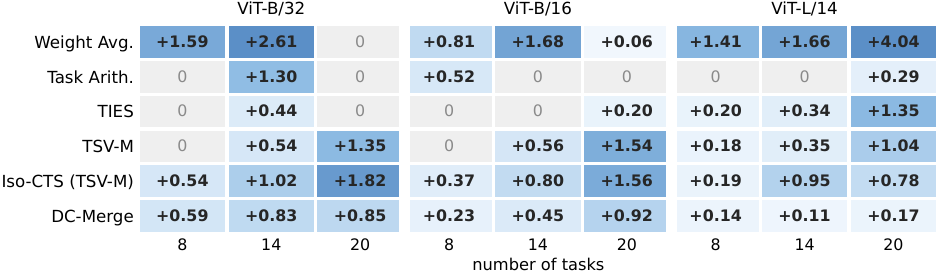}
\caption{Gain of \proj{} over each base method of
Table~\ref{tab:method-agnostic}, in test normalized accuracy. Grey cells select
$\alpha_1^{\star}=0$ and return the base; bold marks a gain.}
\label{fig:gain-heatmap}
\end{figure*}

\section{Per-Task Results with 20 Tasks}
\label{app:pertask-20t}

Table~\ref{tab:pertask-20t-full} reports the complete per-task results for
DC-Merge on the largest suite. The gains are concentrated on tasks with a
large initial deficit. On ViT-B/16, KMNIST gains $+16.5$ absolute points and
SVHN gains $+4.8$. Declines on the remaining tasks are at most $1.4$ points.
The setting gains $+0.90$ points on average. The same pattern appears on
ViT-B/32 and ViT-L/14, with mean gains of $+0.78$ and $+0.16$ points.

\begin{table*}[t]
\centering
\resizebox{\textwidth}{!}{%
\begin{tabular}{l|ccc|ccc|ccc}
\toprule
\textbf{Task} & \multicolumn{3}{c|}{ViT-B/32} & \multicolumn{3}{c|}{ViT-B/16} & \multicolumn{3}{c}{ViT-L/14} \\
\cmidrule(lr){2-4}\cmidrule(lr){5-7}\cmidrule(lr){8-10}
 & Base & +\,HR & $\Delta$ & Base & +\,HR & $\Delta$ & Base & +\,HR & $\Delta$ \\
\midrule
KMNIST & 74.95 & \textbf{77.59} & $+2.6$ & 67.62 & \textbf{84.16} & $+16.5$ & 90.43 & 88.78 & $-1.6$ \\
SVHN & 76.45 & \textbf{91.39} & $+14.9$ & 84.90 & \textbf{89.69} & $+4.8$ & 90.42 & \textbf{95.43} & $+5.0$ \\
Cars & 56.26 & \textbf{59.67} & $+3.4$ & 69.44 & \textbf{70.91} & $+1.5$ & 85.92 & \textbf{86.21} & $+0.3$ \\
CIFAR100 & 69.65 & 69.37 & $-0.3$ & 74.01 & \textbf{74.75} & $+0.7$ & 83.65 & \textbf{84.57} & $+0.9$ \\
SUN397 & 66.91 & \textbf{67.13} & $+0.2$ & 70.32 & \textbf{70.84} & $+0.5$ & 76.38 & \textbf{76.64} & $+0.3$ \\
Food101 & 81.58 & \textbf{81.75} & $+0.2$ & 87.69 & \textbf{88.18} & $+0.5$ & 93.97 & 93.97 & $+0.0$ \\
PCAM & 84.35 & 84.24 & $-0.1$ & 85.03 & \textbf{85.20} & $+0.2$ & 86.22 & \textbf{86.53} & $+0.3$ \\
OxfordIIITPet & 90.54 & \textbf{90.62} & $+0.1$ & 94.00 & \textbf{94.14} & $+0.1$ & 95.78 & \textbf{95.97} & $+0.2$ \\
Flowers102 & 77.31 & 77.00 & $-0.3$ & 83.58 & \textbf{83.64} & $+0.1$ & 92.62 & 91.46 & $-1.2$ \\
RenderedSST2 & 71.22 & \textbf{71.88} & $+0.7$ & 77.32 & \textbf{77.38} & $+0.1$ & 84.02 & \textbf{84.35} & $+0.3$ \\
STL10 & 97.54 & 97.38 & $-0.2$ & 98.16 & 98.09 & $-0.1$ & 99.54 & \textbf{99.55} & $+0.0$ \\
CIFAR10 & 93.89 & 93.48 & $-0.4$ & 95.07 & 94.81 & $-0.3$ & 98.03 & 98.02 & $-0.0$ \\
RESISC45 & 81.44 & 80.97 & $-0.5$ & 85.81 & 85.41 & $-0.4$ & 93.06 & 92.83 & $-0.2$ \\
MNIST & 92.22 & \textbf{93.19} & $+1.0$ & 95.99 & 95.52 & $-0.5$ & 95.64 & \textbf{96.51} & $+0.9$ \\
FER2013 & 64.08 & 63.60 & $-0.5$ & 67.33 & 66.84 & $-0.5$ & 71.78 & 71.59 & $-0.2$ \\
DTD & 66.17 & 66.12 & $-0.1$ & 71.01 & 70.37 & $-0.6$ & 77.77 & \textbf{77.93} & $+0.2$ \\
EuroSAT & 90.70 & 87.70 & $-3.0$ & 96.85 & 96.11 & $-0.7$ & 98.11 & 97.93 & $-0.2$ \\
EMNIST & 97.12 & 95.75 & $-1.4$ & 97.85 & 96.89 & $-1.0$ & 99.50 & 99.36 & $-0.1$ \\
FashionMNIST & 88.07 & 86.87 & $-1.2$ & 91.05 & 89.61 & $-1.4$ & 92.87 & 92.60 & $-0.3$ \\
GTSRB & 82.22 & \textbf{82.64} & $+0.4$ & 89.54 & 88.10 & $-1.4$ & 95.55 & 94.21 & $-1.3$ \\
\midrule
\textit{Cell mean} & & & $+0.78$ & & & $+0.90$ & & & $+0.16$ \\
\bottomrule
\end{tabular}
}
\caption{Absolute per-task accuracy of DC-Merge and \proj{} on the 20-task
suite, sorted by the ViT-B/16 gain.}
\label{tab:pertask-20t-full}
\end{table*}

Across all seven base configurations, including the symmetric Iso-CTS
variant, the 63 full fine-tuning cells give 882 task-level comparisons. The
refinement is active in 41 cells and returns the base model in 22 cells.

\section{Selective Refinement Cases}
\label{app:failure}

Figure~\ref{fig:gain-heatmap} shows the gain of \proj{} over each base in
Table~\ref{tab:method-agnostic}; its grey cells are discussed below.

The refinement is useful only if the residual merge supplies a direction
that improves the validation objective. TIES activates Stage~1 in five of
nine full fine-tuning settings. The other four select
$\alpha_1^{\star}=0$ and return the base model. The active settings are
concentrated at larger task scales.

The symmetric Iso-CTS variant uses Iso-CTS in both stages. It returns the
base model in all nine settings. Iso-CTS equalizes the retained singular
directions, which absorbs the task-dependent residual weights into a global
scale. Replacing only the Stage~1 operator with TSV-M preserves those
differences and activates refinement in all nine settings.
Table~\ref{tab:method-agnostic} reports the TSV-M variant and omits the
symmetric one; the coefficients of both appear in
Table~\ref{tab:hyperparams-full}.

Task arithmetic activates Stage~1 in three of nine settings and returns the
base in the other six, including ViT-B/32 with 20 tasks. Its operator sums
the task vectors with fixed coefficients, so the Stage-1 direction is itself
a combination of the task vectors, and the refined model differs from the
base only by a headroom-dependent rescaling of each task vector. When no such
rescaling raises the validation mean, the selection returns the base.

In the language suite, Core-space TIES concentrates the Stage-1 weight on
QNLI ($w_t=3.07$; Table~\ref{tab:nli-pertask}). MNLI and SciTail, with small
headroom and weights below $0.7$, lose about $0.6$ points each, while QNLI
gains $6.39$ and the average gains $1.00$. As in the vision suites, the
residual weighted toward one task is not neutral for the others
(Section~\ref{sec:motivation}).

\section{DTD Evaluation Integrity}
\label{app:integrity}

The inherited evaluation directory for DTD contains the Oxford VGG
\texttt{val1} partition. The benchmark training set already includes
\texttt{train1} and \texttt{val1}. Evaluating on that directory therefore
reuses training data. Direct evaluation of the same expert gives
$97.71\%$ on \texttt{val1} and $79.20\%$ on the official \texttt{test1}
split. The commonly circulated value near $97.98\%$ matches the former: it
is the DTD entry of the expert-accuracy file released with the Iso-C code
(ViT-B/32; $98.30$ and $98.51$ for B/16 and L/14, against $81.76$ and $84.26$
on \texttt{test1}). TSV-M, Iso-CTS and DC-Merge state that they reuse the
TALL-masks code and checkpoints \citep{gargiulo2025tsv,marczak2025iso,zhang2026dcmerge},
so the DTD column and the averages of their main 8/14/20-task tables follow
this split.

The loader and single-task accuracy file are identical across the inherited
repositories we inspected. A path-overlap check
finds this issue only for DTD among the 20 tasks. We therefore evaluate DTD
on \texttt{test1} throughout. The expert checkpoint needs no retraining
because its training data follow the intended protocol of the benchmark.

Correcting the split changes absolute averages but leaves the method ranking
unchanged. Our paired comparisons use the same corrected split, expert
normalizer, and evaluation pipeline for every base and \proj{} model.

\section{Compute and Budget Accounting}
\label{app:budget}

All experiments run on single-node consumer GPUs. RTX~4090 nodes are used
for the main vision study and the motivating experiments. A two-GPU RTX~5090
node is used for the temperature-recalibration and redeployment batch and for
both LoRA matrices. The vision runs use PyTorch~2.8.0 (CUDA~12.8),
torchvision~0.23.0, Pillow~11.3.0, and open-clip~2.10. The LLaMA-3 language
runs use the same PyTorch build under Python~3.9, with transformers~4.43.1,
peft~0.15.2, accelerate~1.6.0, datasets~2.16.1, safetensors~0.5.3,
sentencepiece~0.1.97, numpy~1.26.3, and scipy~1.11.4. The merges are seeded
with 420 and deterministic, and we archive a bit-exact confirmation run.
Our code and the environment exports for both stacks are included as
supplementary material.
The wrapper adds no training, so its budget is merging plus evaluation, and
evaluation dominates.

The cost model is given in Appendix~\ref{app:complexity}. Here we report the
deployed budget in evaluations rather than wall-clock time, making it
hardware-independent. One evaluation is a single forward pass over all tasks
in one cell. One task-pass is the corresponding pass on one task, so a cell
with $T$ tasks costs $T$ task-passes per evaluation.

Expert validation accuracies are precomputed on each task's validation split
and reused across merges. The headroom scores are read from the validation
pass already used by the base protocol to select $\alpha_0$ and therefore add no evaluation. Per cell,
the wrapper adds the $\alpha_1$ validation sweep and one test evaluation of
the selected model:

\begin{center}
\centering
\small
\setlength{\tabcolsep}{4pt}
\begin{tabular}{lcl}
\toprule
Suite & Grid & Added per cell \\
\midrule
Vision, full fine-tuning & 21 points & $\leq$\,21 val. $+$ 1 test \\
Vision LoRA              & 7 points  & 7 val. $+$ 1 test \\
Language LoRA            & 12 points & $\leq$\,12 val. $+$ 1 test \\
\bottomrule
\end{tabular}
\end{center}

The full fine-tuning grid is $\{0,0.05,\ldots,1\}$. The vision LoRA suite
uses the coarser grid $\{0,0.1,0.2,\allowbreak 0.3,0.4,\allowbreak 0.6,1\}$, extended above $1$ for
TIES+Iso-C in Full space (Appendix~\ref{app:hyperparams}). The language suite
uses the finer but shorter grid
$\{0,0.01,0.02,0.03,0.05,0.08,\allowbreak
0.1,0.15,0.2,0.3,0.4,0.5\}$, which resolves
the small strengths selected on that benchmark. Each sweep visits its grid in
increasing order. Under full fine-tuning, the sweeps for DC-Merge, weight
averaging, and both Iso-CTS variants stop after three consecutive points that
do not improve on the best validation objective so far, as do the language
sweeps once $\alpha_1\geq0.1$; task arithmetic, TIES, TSV-M, and the vision
LoRA cells scan their full grids. Because $\alpha_1=0$ is evaluated first,
stopping early leaves Proposition~\ref{prop:failclosed} intact. Summed over
all studies, the wrapper adds at most $19{,}954$ validation task-passes and
$1{,}038$ test task-passes: at most $18{,}522$ and $882$ for the 63 full
fine-tuning cells, $712$ and $96$ for the twelve vision LoRA cells (including
five extension points), and at most $720$ and $60$ for the ten language
cells.

Two costs are deliberately excluded. Calibrating the temperature constant
$c$ is not part of the deployed budget. It is performed once per base method
on the ViT-B/32 8-task cell and then frozen for every other backbone, task
count, and suite, so no deployment cell pays this cost. It takes
$30\times11=330$ validation evaluations of the 8-task ViT-B/32 model, about
5--6 GPU-hours on one RTX 4090, plus about 8 GPU-hours for the 14-task
fallback of TIES. The base merger's own
$\alpha_0$ selection is also excluded because each method runs its published
protocol whether or not it is wrapped.

\section{Computational Complexity}
\label{app:complexity}

\proj{} is a wrapper, so its cost is stated relative to the base merger it
wraps. Let $T$ denote the number of tasks, $P$ the parameter count of one
model, $C_{\mathcal{M}}$ the cost of one application of the base operator
$\mathcal{M}$, and $C_{\mathrm{eval}}$ the cost of one evaluation pass over the $T$
validation sets; the wrapper adds no training.

\paragraph{Merging.}
Stage~0 computes
$\Delta_{\mathrm{M}}^0=\mathcal{M}(\{\Delta_t\})$, and Stage~1 computes
$\Delta_{\mathrm{M}}^1=\mathcal{M}(\{w_t\varepsilon_t\})$
(Algorithm~\ref{alg:hr-merge}). The operator is therefore invoked twice
instead of once, on the same number of inputs with the same shape. The merge
cost is $2C_{\mathcal{M}}$, compared with the base cost
$C_{\mathcal{M}}$. For SVD-based operators, this doubles the SVD count. On
the LLaMA-3 8B benchmark, a Full-space TSV or DARE merge performs 548
layer-wise full-dimensional SVDs and takes roughly 22 minutes. This is a
floor imposed by the decomposition itself. Thus,
\proj{} takes about 44 minutes. In Core space, the same merge runs in seconds,
making the doubling negligible.

\paragraph{Weighting.}
The headroom vector $\mathbf{g}$ is read from the validation pass already used
by the base protocol to select $\alpha_0$, so it incurs no additional
evaluation. Everything downstream is $O(T)$ scalar work: one pass computes
$\operatorname{std}(\mathbf{g})$, and another computes the softmax
(Eq.~\ref{eq:headroom-weights}). With $P$ in the hundreds of millions and $T\leq20$, this term is
negligible.

\paragraph{Searching.}
The $\alpha_1$ grid does not re-merge. Both
$\Delta_{\mathrm{M}}^0$ and $\Delta_{\mathrm{M}}^1$ are computed once. Each
grid point forms the scalar combination
$\theta_0+\alpha_0\Delta_{\mathrm{M}}^0+
\alpha_1\Delta_{\mathrm{M}}^1$, requiring $O(P)$ arithmetic followed by one
evaluation. The full fine-tuning search therefore costs at most
$21C_{\mathrm{eval}}$ and no additional merges. Its width affects only the
number of evaluations, not the number of merges, and this cost is independent
of backbone, task count, and operator
(Section~\ref{sec:strength-selection}).

\paragraph{What dominates in practice.}
Evaluation dominates, and it is essentially the entire cost when the merge
operator is cheap. On ViT-B/16 with one RTX~4090 and no feature cache (the
validation activations need 12.3\,GB against 15\,GB of host memory), one
$\alpha_1$ evaluation takes 8.6, 12.2, and 16.6 minutes for 8, 14, and 20
tasks, so a full 21-point sweep takes roughly 3--6 hours per cell before the
early stop. The base protocol runs an $\alpha_0$ sweep of the same order on
the same harness.

\paragraph{Memory.}
Stage~1 requires $\Delta_{\mathrm{M}}^0$ and
$\Delta_{\mathrm{M}}^1$ to be resident simultaneously. Peak merge memory is
therefore one model larger than for the base method. The reweighted residuals
$w_t\varepsilon_t$ are formed in place. The residuals need not be stored:
$\varepsilon_t$ for one weight tensor needs only that tensor of $\Delta_t$ and
$\Delta_{\mathrm{M}}^0$, so an operator that acts tensor by tensor can form,
weight and merge the residuals one tensor at a time, and the wrapper adds one
model-sized buffer to what the base merge already reads. Our ViT-L/14 runs
with 20 tasks merge this way, one tensor at a time from disk, on a host with
15\,GB of memory.

\paragraph{Summary.}
\proj{} doubles the merge computation, adds at most 21 validation evaluations
per full fine-tuning cell and one model of memory, and changes no asymptotic
term of the base method.

\section{Per-Example Significance}
\label{app:significance}

\begin{table}[ht]
\centering
\small
\setlength{\tabcolsep}{3pt}
\resizebox{\columnwidth}{!}{%
\begin{tabular}{lrrrrrr}
\toprule
Setting & $n$ & $b$ & $c$ & $p$ & $\hat\rho$ & w/o SVHN \\
\midrule
B/32, 8  & 87{,}433  & 2620 & 1022 & $10^{-159}$ & 4.2\% & $0.063$ \\
B/32, 14 & 155{,}197 & 3997 & 1461 & $10^{-268}$ & 3.5\% & $8\!\times\!10^{-6}$ \\
B/32, 20 & 252{,}268 & 7117 & 3386 & $10^{-296}$ & 4.2\% & $0.048^{\dagger}$ \\
B/16, 8  & 87{,}433  & 916  & 367  & $10^{-54}$  & 1.5\% & $0.85^{\dagger}$ \\
B/16, 14 & 155{,}197 & 2242 & 933  & $10^{-123}$ & 2.0\% & $10^{-37}$ \\
B/16, 20 & 252{,}268 & 5033 & 2533 & $10^{-185}$ & 3.0\% & $10^{-58}$ \\
L/14, 8  & 87{,}433  & 704  & 365  & $10^{-25}$  & 1.2\% & $0.34^{\dagger}$ \\
L/14, 14 & 155{,}197 & 580  & 329  & $7\!\times\!10^{-17}$ & 0.6\% & $5\!\times\!10^{-4}$ \\
L/14, 20 & 252{,}268 & 2570 & 1415 & $10^{-75}$  & 1.6\% & $0.004^{\dagger}$ \\
\bottomrule
\end{tabular}}
\caption{Exact two-sided McNemar test of \proj{} against its DC-Merge base,
pooled over the test images of each setting. $b$ counts images only \proj{}
classifies correctly, $c$ those only the base does, and
$\hat\rho=(b+c)/n$ is the test disagreement rate. The last column gives the
$p$-value without SVHN. $^{\dagger}$Here $c>b$.}
\label{tab:mcnemar}
\end{table}

Section~\ref{sec:vision-full-finetuning} counts settings: all nine changes
are positive. A per-example test has far more units. We store, for every test
image, whether the base and \proj{} classify it correctly, and test equal
accuracy with an exact two-sided McNemar test. Only the $b+c$ images on
which the two models disagree carry information, and each setting has
between about one and ten thousand of them.

Table~\ref{tab:mcnemar} reports nine DC-Merge settings at the deployed
$\alpha_1^{\star}$. \proj{} wins $64$--$73\%$ of the disagreements, and every
setting rejects equal accuracy. The net gain is concentrated: SVHN contributes
the largest share in eight settings and KMNIST in the ninth (ViT-B/16 with 20
tasks). Without SVHN, the three eight-task settings are no longer
significant, and the 20-task settings on ViT-B/32 and ViT-L/14 lose more
images than they gain ($p=0.048$ and $0.004$); the base also wins more of the remaining disagreements on
ViT-B/16 and ViT-L/14 with 8 tasks. Per task, significant losses fall mostly on tasks close to
their ceiling, such as MNIST, EuroSAT, GTSRB and EMNIST, while the significant
gains fall on SVHN, Cars, SUN397, CIFAR100 and, on ViT-B/16, KMNIST, which
declines on ViT-L/14 with 20 tasks. Pooling weights tasks
by test-set size, whereas Table~\ref{tab:method-agnostic} weights them
equally. Table~\ref{tab:tasklevel} therefore also tests the reported metric
itself, with a task-stratified paired interval for the mean normalized change.
It rejects no change in all nine settings. Without SVHN the mean change stays
significantly positive in four of them and is negative on ViT-L/14 with 20
tasks ($-0.09\pm0.07$).
Per task, improvements and declines are close to balanced: \proj{} moves
accuracy toward the tasks the merge hurt most, and a few large recoveries
carry the mean.

\begin{table*}[t]
\centering
\footnotesize
\setlength{\tabcolsep}{4pt}
\begin{tabularx}{\textwidth}{@{}l>{\centering\arraybackslash\hsize=0.85\hsize}X>{\centering\arraybackslash\hsize=0.85\hsize}X|>{\centering\arraybackslash\hsize=1.4\hsize}X>{\centering\arraybackslash\hsize=0.75\hsize}X|>{\centering\arraybackslash\hsize=1.4\hsize}X>{\centering\arraybackslash\hsize=0.75\hsize}X@{}}
\toprule
 & & & \multicolumn{2}{c|}{All tasks} & \multicolumn{2}{c}{Without SVHN} \\
Setting & Tasks $\uparrow$/$\downarrow$ & Median $\Delta_t$ & $\Delta$ & $p$ & $\Delta$ & $p$ \\
\midrule
B/32, 8  & 3/5   & $-0.13$ & $+0.57\pm0.20$ & $<10^{-4}$ & $-0.20\pm0.23$ & $0.085$ \\
B/32, 14 & 5/9   & $-0.08$ & $+0.83\pm0.12$ & $<10^{-4}$ & $+0.20\pm0.13$ & $0.002$ \\
B/32, 20 & 9/11  & $-0.09$ & $+0.85\pm0.12$ & $<10^{-4}$ & $+0.09\pm0.13$ & $0.18$ \\
B/16, 8  & 3/5   & $-0.08$ & $+0.23\pm0.11$ & $<10^{-4}$ & $-0.05\pm0.12$ & $0.43$ \\
B/16, 14 & 7/7   & $+0.01$ & $+0.45\pm0.10$ & $<10^{-4}$ & $+0.28\pm0.10$ & $<10^{-4}$ \\
B/16, 20 & 10/10 & $-0.00$ & $+0.92\pm0.10$ & $<10^{-4}$ & $+0.71\pm0.11$ & $<10^{-4}$ \\
L/14, 8  & 4/4   & $+0.05$ & $+0.14\pm0.10$ & $0.007$    & $-0.04\pm0.11$ & $0.48$ \\
L/14, 14 & 10/3  & $+0.05$ & $+0.11\pm0.05$ & $<10^{-4}$ & $+0.07\pm0.06$ & $0.012$ \\
L/14, 20 & 11/8  & $+0.01$ & $+0.17\pm0.07$ & $<10^{-4}$ & $-0.09\pm0.07$ & $0.015$ \\
\bottomrule
\end{tabularx}
\caption{Task-level view of the same runs: tasks improved/declined, median
per-task change, and mean normalized change (equal task weights, as in
Table~\ref{tab:method-agnostic}) with a task-stratified $95\%$ interval
and $p$-value.}
\label{tab:tasklevel}
\end{table*} The disagreement rates here are measured on test images;
Appendix~\ref{app:slack} gives the validation rates that
Theorem~\ref{thm:transfer} uses.

These runs use a different GPU from the main experiments. The base models
reproduce Table~\ref{tab:method-agnostic} exactly in all nine settings, and
the refined models agree within $0.06$ points on every task.

\paragraph{The language study.}
We re-run Core-space TIES on the NLI benchmark at the deployed $\alpha_0=1.1$
and $\alpha_1^{\star}=0.4$ on the same GPU. It gives $93.28\to94.26$, against
$93.29\to94.29$ in Table~\ref{tab:lora-nli-matrix}, with every task within
$0.24$ points. On the $29{,}216$ test pairs, \proj{} is right on $574$ of the
$897$ pairs where the two models disagree ($p=4\times10^{-17}$), and the
task-stratified change of the mean is $+0.98\pm0.21$ ($p<10^{-4}$). QNLI
carries the gain ($260$ against $7$). Without it the remaining disagreements
split $314$ to $316$ ($p=0.97$), and the mean change is $-0.05\pm0.21$. Two
controls repeat Table~\ref{tab:ladder} on this cell with the same grid and
selection rule. Uniform Stage-1 weights gain $+0.01$. Three permutations of
the \proj{} weights, each moving QNLI off the largest weight, gain
$+0.00$, $+0.00$ and $+0.03$, against $+0.98$ for \proj{}.

\section{Ablation Details}
\label{app:ablation-details}

\paragraph{Permuted weights.}
Each permutation reassigns the \proj{} weight vector across tasks and is
resampled until the most-weighted task moves. The temperature is unchanged,
because $\operatorname{std}(g)$ does not depend on the order of the tasks.
Strength selection follows the deployed protocol
(Table~\ref{tab:permutation-seeds}).

\paragraph{Summed and single-stage variants.}
Two further controls keep the base merge, the \proj{} weights and the
selection budget, and change only where the weights act
(Table~\ref{tab:permutation-seeds}). The summed variant adds the weighted
residuals instead of re-merging them. A sum of $T$ residuals is on a larger
scale than a merge, so its strength grid is $\{0,0.05,\dots,1\}$ times the
ratio of the task-arithmetic and DC-Merge base scales
(Table~\ref{tab:hyperparams-full}). In all three settings its validation
accuracy falls from the first non-zero strength, so it selects
$\alpha_1=0$ and returns the base. The single-stage variant multiplies the
task vectors by the same weights inside the Stage-0 merge and has no
Stage~1; $\alpha_0$ is re-selected on validation over
$\{1.0,1.1,\dots,3.0\}$, the same 21-point budget. Its best models lose
$6.58$ and $4.39$ normalized points on test with 8 and 14 tasks. With 20
tasks the weighted Stage-0 merge does not fit in the memory of our
evaluation host, so we do not report it. The weights therefore help only as
a correction on top of an unweighted merge, and only when the weighted
residuals are merged rather than summed.

\begin{table}[ht]
\centering
\footnotesize
\setlength{\tabcolsep}{4pt}
\begin{tabular}{lccc}
\toprule
 & 8 tasks & 14 tasks & 20 tasks \\
\midrule
seed 0 & $+0.10$ (0.10) & $+0.00$ (0) & $+0.00$ (0) \\
seed 1 & $+0.00$ (0) & $+0.29$ (0.30) & $+0.00$ (0) \\
seed 2 & $+0.01$ (0.05) & $+0.00$ (0) & $+0.06$ (0.05) \\
\midrule
summed & $+0.00$ (0) & $+0.00$ (0) & $+0.00$ (0) \\
single stage & $-6.58$ (1.4) & $-4.39$ (1.5) & -- \\
\midrule
\proj{} & $+0.23$ (0.25) & $+0.45$ (0.35) & $+0.92$ (0.45) \\
\bottomrule
\end{tabular}
\caption{Stage-1 controls on ViT-B/16: test gain in normalized accuracy over
the DC-Merge base, with the selected $\alpha_1^{\star}$ in parentheses (for
the single-stage variant, the re-selected $\alpha_0$). The seeds permute the
\proj{} weights across tasks; the other two controls are described in the
text.}
\label{tab:permutation-seeds}
\end{table}

\paragraph{Base-scale retuning.}
With Stage~1 disabled, the ViT-B/32 validation curve has a single peak at the
deployed $\alpha_0=1.90$ ($95.72$, against $94.48$ at $1.40$ and $93.92$ at
$2.40$). The ViT-B/16 curve is flat near the top ($96.88$ at $1.90$, $96.89$
at $1.95$), and the pick of $1.95$ scores $95.78$ on test against $95.87$ for
the deployed base and $96.10$ for \proj{}.

\paragraph{Validation budget.}
\proj{} uses labeled validation data twice, for the headroom and for the
strength. We keep a random fraction of every validation split for both,
select as deployed, and evaluate once on the full test split; $\alpha_0$
stays at its deployed value, selected on all validation data
(Table~\ref{tab:valfrac}). No run falls below the base. With half of the
validation data, every run gains at least $+0.62$ points and the mean is
within $0.05$ points of the full-data gain, or above it, on all three suites. With a quarter,
the mean matches or exceeds the full-data gain on 8 and 20 tasks but keeps
$60\%$ of it on 14 tasks: there, the early stop of
Appendix~\ref{app:budget} ends all three sweeps after seven or eight points,
before the full-data optimum $\alpha_1^{\star}=0.40$, on a noisier validation
curve. With a tenth, as few as 38 images for DTD, the gain shrinks and
varies, and one subsample on 8 tasks returns the base. The search cost
scales with the validation size: on 8 tasks and one RTX~4090, one $\alpha_1$
evaluation takes $64.8$, $29.2$, and $14.8$\,s with all, half, and a quarter
of the data.

\begin{table}[ht]
\centering
\footnotesize
\setlength{\tabcolsep}{4pt}
\resizebox{\columnwidth}{!}{%
\begin{tabular}{lcccc}
\toprule
Tasks & 10\% & 25\% & 50\% & 100\% \\
\midrule
8  & $+0.31$ ($+0.00$) & $+0.74$ ($+0.55$) & $+0.70$ ($+0.62$) & $+0.59$ \\
14 & $+0.45$ ($+0.14$) & $+0.49$ ($+0.33$) & $+0.78$ ($+0.73$) & $+0.83$ \\
20 & $+0.60$ ($+0.35$) & $+0.89$ ($+0.84$) & $+0.84$ ($+0.76$) & $+0.85$ \\
\bottomrule
\end{tabular}}
\caption{Validation budget on ViT-B/32: test gain in normalized accuracy over
the DC-Merge base when the headroom and $\alpha_1$ use a random fraction of
every validation split. Each entry is the mean over three subsamples, with
the smallest in parentheses; the 100\% column is the deployed run, which
uses no subsample.}
\label{tab:valfrac}
\end{table}

\section{Task-Level Effects}
\label{app:task-level}

\subsection{Imbalance Across Tasks}
\label{app:imbalance}

\begin{table*}[t]
\centering
\small
\setlength{\tabcolsep}{4pt}
\begin{tabularx}{\textwidth}{@{}lYY|YY|YY@{}}
\toprule
 & \multicolumn{2}{c|}{Std.\ across tasks ($\downarrow$)} & \multicolumn{2}{c|}{Worst task ($\uparrow$)} & \multicolumn{2}{c}{Bottom quartile ($\uparrow$)} \\
Setting & Base & +\,HR & Base & +\,HR & Base & +\,HR \\
\midrule
DC-Merge, B/32, 8 tasks & 3.36 & \textbf{2.81} & 87.96 & \textbf{89.58} & 89.21 & \textbf{90.43} \\
DC-Merge, B/32, 14 tasks & 6.43 & \textbf{5.47} & 77.99 & \textbf{81.88} & 82.91 & \textbf{85.33} \\
DC-Merge, B/32, 20 tasks & 7.91 & \textbf{6.94} & 71.96 & \textbf{76.32} & 77.45 & \textbf{79.90} \\
DC-Merge, B/16, 8 tasks & 2.19 & \textbf{1.88} & 92.84 & \textbf{93.37} & 93.36 & \textbf{93.91} \\
DC-Merge, B/16, 14 tasks & 4.20 & \textbf{3.78} & 85.88 & \textbf{88.20} & 88.51 & \textbf{89.54} \\
DC-Merge, B/16, 20 tasks & 7.63 & \textbf{5.42} & 68.45 & \textbf{80.94} & 80.59 & \textbf{84.54} \\
DC-Merge, L/14, 8 tasks & 1.18 & \textbf{0.92} & 96.61 & \textbf{96.80} & 96.68 & \textbf{96.99} \\
DC-Merge, L/14, 14 tasks & 2.39 & \textbf{2.24} & 92.24 & \textbf{93.28} & 93.92 & \textbf{94.16} \\
DC-Merge, L/14, 20 tasks & 3.06 & \textbf{3.04} & 89.63 & \textbf{89.90} & 91.74 & \textbf{91.87} \\
\midrule
LLaMA-3, TIES (Core) & 5.00 & \textbf{3.09} & 83.63 & \textbf{90.02} & 87.88 & \textbf{91.07} \\
ViT-B/32 LoRA, TSV+Iso-C (Core) & 15.68 & \textbf{13.38} & 50.88 & \textbf{53.62} & 52.23 & \textbf{57.22} \\
ViT-B/32 LoRA, CART+Iso-C (Full) & 13.61 & 13.72 & 51.79 & \textbf{55.00} & 55.17 & \textbf{55.59} \\
\bottomrule
\end{tabularx}
\caption{Imbalance of per-task test normalized accuracy before and after
\proj{}. The bottom quartile averages the worst $\lceil T/4\rceil$ tasks of
each model. Bold marks an improvement. None of these statistics enters the
strength selection, which maximizes the mean.}
\label{tab:imbalance}
\end{table*}

The strength selection maximizes validation accuracy normalized by expert
validation accuracy, so a
gain in the mean could in principle come from moving accuracy between tasks rather than
from lifting the tasks the base merge harmed. Table~\ref{tab:imbalance}
separates the two. Across twelve settings, \proj{} raises the worst task in
all twelve and the mean of the bottom quartile in all twelve, and it lowers
the standard deviation across tasks in eleven. The exception, CART+Iso-C in
Full space, rises by $0.11$. A uniform shift of every task would leave the
standard deviation unchanged, so its consistent decrease shows that the gains
fall disproportionately on the weakest tasks.

\subsection{Headroom and Recoverability}
\label{app:recoverability}

\begin{figure}[t]
\centering
\includegraphics[width=\columnwidth]{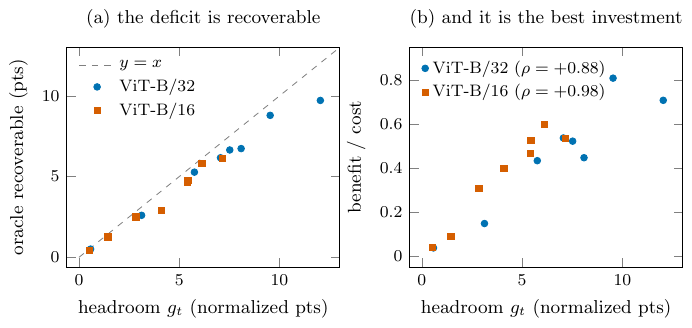}
\caption{Headroom against recoverability on DC-Merge with eight tasks, one
point per task. (a)~Accuracy recovered by adding back the task's top-$10\%$
residual alone; the dashed line is full recovery. (b)~Gain of the same patch
on its own task divided by the mean damage it causes to the other tasks.}
\label{fig:recoverability}
\end{figure}

Headroom measures how much a task has lost. Refinement needs the loss to be
recoverable, and a task could have a large deficit precisely because its
update conflicts with the others. Figure~\ref{fig:recoverability} tests this
with two measurements that involve no headroom weighting: the oracle patch of
Figure~\ref{fig:motivation}a and the interference matrix of
Figure~\ref{fig:motivation}b. On ViT-B/32 and ViT-B/16, adding back each
task's top-$10\%$ residual recovers $71$--$95\%$ of its test deficit. The
recovered amount rises with the deficit (Spearman $\rho=1.00$ on both
backbones), and so does the ratio of a patch's gain on its own task to its
mean damage on the others ($\rho=0.88$ and $0.98$, exact $p<0.01$, $T=8$).
The tasks with the largest deficit are therefore the most recoverable, not
the hardest. With the deployed validation headroom in place of the test
deficit, the same correlations remain positive but weaker
($\rho=0.31$--$0.62$) and are not significant with eight tasks.

Figure~\ref{fig:motivation}c correlates headroom with the gain of a model
whose Stage-1 weights are themselves derived from headroom. The uniform row of
Table~\ref{tab:ladder} provides the control on ViT-B/16. With $w_t=1$, the
Spearman correlation between validation headroom and the per-task normalized
test gain is $-0.29$, $-0.09$, and $+0.13$ with 8, 14, and 20 tasks
(two-sided permutation test, $p\ge0.50$). With the headroom weights, in the
same ablation run, it is $+0.21$, $+0.21$, and $+0.43$; with at most twenty
tasks per setting, none of these is significant on its own either. Headroom
therefore does not predict by itself which tasks gain. It decides where the
refinement goes, and the permuted row of Table~\ref{tab:ladder}, which keeps
the weight values but reassigns them across tasks, shows that the gain
comes from this allocation rather than from the weight values.

\subsection{A Multi-Task Training Reference}
\label{app:mtl-reference}

\begin{table}[ht]
\centering
\small
\setlength{\tabcolsep}{4pt}
\begin{tabular}{lcccc}
\toprule
 & \multicolumn{2}{c}{Reported} & \multicolumn{2}{c}{Ours} \\
\cmidrule(lr){2-3}\cmidrule(lr){4-5}
Eight tasks & Experts & Joint MTL & Experts & \proj{} \\
\midrule
ViT-B/32 & 90.5 & 88.9 & 90.51 & 85.21 \\
ViT-L/14 & 94.2 & 93.5 & 94.10 & 92.55 \\
\bottomrule
\end{tabular}
\caption{Average absolute test accuracy of individual experts and of one model
trained jointly on all eight tasks, as reported by \citet{yang2024adamerging},
next to our experts and the DC-Merge variant of \proj{}.}
\label{tab:mtl-reference}
\end{table}

Normalized accuracy measures a merged model against its individual experts,
which no single model need reach. A model trained jointly on all eight tasks
gives a second reference. Table~\ref{tab:mtl-reference} takes it from
\citet{yang2024adamerging}, whose experts score within $0.1$ points of ours on
average; the joint model is their own training run, so the comparison is
indicative only. On ViT-L/14, the joint model trails the experts by $0.7$
points and \proj{} comes within $1.0$ point of it, so little multi-task
accuracy remains to recover. On ViT-B/32, \proj{} remains $3.7$ points below
the joint model, which matches the larger gains we observe there.

\section{Theoretical Details}
\label{app:theory}

This appendix completes Section~\ref{sec:theory}. It defines the candidate
set, states the assumptions, proves Theorem~\ref{thm:transfer}, reports the
measured slack, and shows that one round suffices for fixed-coefficient
operators.

\subsection{Candidate Set}
\label{app:candidate-set}

The selection ranges over
\begin{equation}
\mathcal{C} = \mathcal{A}_0\times\mathcal{T}\times
\left(\{0\}\cup\mathcal{G}\right),
\label{eq:candidate-set}
\end{equation}
where $\mathcal{A}_0$ is the base method's own grid for $\alpha_0$,
$\mathcal{T}=\{0.1,0.2,\ldots,3.0\}$ is the calibration sweep for the
temperature constant $c$ (Appendix~\ref{app:hyperparams}), and
$\{0\}\cup\mathcal{G}=\{0,0.05,\ldots,1.0\}$ has $21$ points. We count
$\mathcal{T}$ in every setting, although $c$ is calibrated once per base
method and then frozen, because the calibration reads the same validation
tasks. The base-scale grid $\mathcal{A}_0$ has $30$ points for every base
(Appendix~\ref{app:hyperparams}), so $|\mathcal{C}|=30\cdot30\cdot21=18{,}900$,
the value used in Appendix~\ref{app:slack}.
Since $\varepsilon$ grows with $L=\ln(2|\mathcal{C}|/\delta)$, a tenfold
larger $\mathcal{C}$ would raise $\varepsilon$ by less than $20\%$.

\subsection{Assumptions}
\label{app:assumptions}

\begin{assumption}[Exchangeable splits]
\label{as:iid}
For each task, the validation and test examples are drawn independently from
the same distribution.
\end{assumption}

\begin{assumption}[Independent selection split]
\label{as:split}
The Stage-1 weights $w$ come from a validation split disjoint from the one
that selects $\alpha_1$.
\end{assumption}

Assumption~\ref{as:iid} is the usual exchangeability of the released splits.
It holds only approximately for benchmarks whose validation split is carved
from the training data while the test split is released separately, and it
fails where a benchmark's own partition is contaminated, which is why
Appendix~\ref{app:integrity} evaluates DTD on the uncontaminated partition.
Assumption~\ref{as:split} is the one the frozen protocol does not satisfy.
With a single split, $\Delta_{\mathrm{M}}^1$ depends on the examples that
later score it, the candidate models are no longer fixed, and the union bound
below does not apply as written. Splitting the validation data once, with the
weights on one half and the selection on the other, restores the theorem
verbatim at the cost of halving $n$ in each role.

\subsection{Proof of Theorem~\ref{thm:transfer}}
\label{app:transfer}

Fix the selection split and write $\theta_\alpha=\theta_0+\alpha_0
\Delta_{\mathrm{M}}^0+\alpha\Delta_{\mathrm{M}}^1$ for the candidate at
strength $\alpha$. By Assumption~\ref{as:split} the weights $w$, and
hence the direction $\Delta_{\mathrm{M}}^1$, are measurable with respect to a
split independent of the selection split; conditioning on that split makes
$\{\theta_\alpha:\alpha\in\mathcal{C}\}$ a fixed finite family of
$|\mathcal{C}|$ models. All statements below are conditional on it.

\paragraph{The paired difference.}
For an example $x$ with label $y$ let
\begin{align}
D_\alpha(x,y) &= \mathbf{1}\!\left[\theta_\alpha\text{ correct}\right]
 - \mathbf{1}\!\left[\theta^0\text{ correct}\right],
\label{eq:paired-diff}\\
D_\alpha(x,y) &\in \{-1,0,+1\}. \nonumber
\end{align}
Two facts drive the bound. First, $A(\theta_\alpha)-A(\theta^0)
=\mathbb{E}[D_\alpha]$, so bounding the difference never requires bounding
either accuracy. Second, $D_\alpha(x,y)=0$ whenever the two models return the
same prediction on $x$, so
\begin{equation}
\mathbb{E}\!\left[D_\alpha^2\right]
\;\le\;\Pr\!\left[\theta_\alpha(x)\neq\theta^0(x)\right]
\;=\;\rho_\alpha ,
\label{eq:var-bound}
\end{equation}
and the unbiased empirical variance obeys it up to a factor $n/(n-1)$,
with $\hat\rho_\alpha$ in place of $\rho_\alpha$.

\paragraph{Concentration.}
Apply the empirical Bernstein inequality of \citet{maurer2009empirical} to
$(D_\alpha+1)/2\in[0,1]$ over the $n$ selection examples, at confidence
$\delta/|\mathcal{C}|$, and take a union bound over the
$|\mathcal{C}|$ candidates. Undoing the rescaling, with probability at least
$1-\delta$, simultaneously for every $\alpha\in\mathcal{C}$,
\begin{equation}
\mathbb{E}[D_\alpha]\;\ge\;\widehat{\mathbb{E}}[D_\alpha]
-\sqrt{\frac{2\hat{V}_\alpha L}{n}}-\frac{14L}{3(n-1)},
\end{equation}
with $L=\ln(2|\mathcal{C}|/\delta)$ and $\hat V_\alpha$ the empirical
variance of $D_\alpha$. The bound holds in particular at the selected
$\alpha_1^\star$, which is what the union bound buys: $\alpha_1^\star$ is
chosen using the same examples, so no single-candidate bound would apply to
it.

\paragraph{Conclusion.}
By Proposition~\ref{prop:failclosed} applied to the empirical objective,
$\widehat{\mathbb{E}}[D_{\alpha_1^\star}]\ge0$. Substituting and bounding
$\hat V_{\alpha_1^\star}\le\frac{n}{n-1}\hat\rho$ (Eq.~\ref{eq:var-bound}) gives
Eq.~\ref{eq:transfer}. \hfill$\square$

\paragraph{The objective over tasks.}
Theorem~\ref{thm:transfer} is stated for one distribution. The deployed
criterion (Eq.~\ref{eq:refinement-strength}) is a mean over $T$ tasks with
weights $1/a_t$, where $a_t$ is expert validation accuracy, and each task
has $n_t$ selection examples. For the independent-split variant, these
normalizers are measured on the construction split and held fixed during selection.
Applying the theorem to every task at confidence $\delta/T$, i.e.\ with
$L_T=\ln(2T|\mathcal{C}|/\delta)$, and averaging gives, with probability at
least $1-\delta$, $A(\theta^\star)\ge A(\theta^0)-\frac{1}{T}\sum_t\varepsilon_t/a_t$,
where $\varepsilon_t$ is Eq.~\ref{eq:slack} on task $t$ with $L_T$. A single
pooled bound would need identically distributed examples, which a mean over
tasks does not have, and its range term would be set by the smallest task.

\subsection{Measured Slack}
\label{app:slack}

Table~\ref{tab:slack} evaluates the bound over tasks of
Appendix~\ref{app:transfer} on the validation splits that select $\alpha_1$,
with $|\mathcal{C}|=18{,}900$ and $\delta=0.05$. For each task we use the
unbiased variance of the paired difference between the DC-Merge base and
\proj{} at the deployed $\alpha_1^\star$; $\hat\rho$ pools the fraction of
examples on which exactly one of the two is correct, which never exceeds the
disagreement rate of Eq.~\ref{eq:disagreement}. The column
$\varepsilon_{\mathrm{norm}}$ uses expert validation accuracies $a_t$
for the same normalization as the selection criterion. The stored predictions
are fixed for this diagnostic; no model is reselected. The smallest validation splits dominate
(Flowers102 has 102 images), and the slack exceeds every gain in
Table~\ref{tab:method-agnostic}. Our runs violate
Assumption~\ref{as:split}, so the table only illustrates the bound and
certifies no reported model.

\begin{table}[ht]
\centering
\small
\begin{tabular}{lrrrr}
\toprule
Setting & $\min_t n_t$ & $\hat\rho$ & $\varepsilon$ & $\varepsilon_{\mathrm{norm}}$ \\
\midrule
ViT-B/32{,} 8 tasks  & 376 & 4.2\% & 8.14 & 9.79 \\
ViT-B/32{,} 14 tasks & 102 & 2.5\% & 14.27 & 15.81 \\
ViT-B/32{,} 20 tasks & 102 & 2.2\% & 11.81 & 13.28 \\
ViT-B/16{,} 8 tasks  & 376 & 0.9\% & 6.83 & 7.87 \\
ViT-B/16{,} 14 tasks & 102 & 1.1\% & 13.51 & 14.55 \\
ViT-B/16{,} 20 tasks & 102 & 1.9\% & 11.39 & 12.40 \\
ViT-L/14{,} 8 tasks  & 376 & 1.6\% & 6.93 & 7.73 \\
\bottomrule
\end{tabular}
\caption{Slack of the bound over tasks on our validation splits:
$\varepsilon$ in accuracy points, $\varepsilon_{\mathrm{norm}}$ in normalized points.}
\label{tab:slack}
\end{table}

\subsection{One-Round Termination}
\label{app:oneround}

Proposition~\ref{prop:failclosed} makes iterating safe; for fixed-coefficient
base methods it is also unnecessary.

\begin{proposition}[One-round termination]
\label{prop:oneround}
Let $\mathcal{M}\!\left(\{x_t\}_{t=1}^{T}\right)=\sum_t c_tx_t$ with
coefficients $c_t$ that do not depend on the arguments, hold the Stage-1
weights fixed across rounds, and let every strength the second round can
reach lie in $\mathcal{G}$. Then $\alpha_2^\star=0$ and the second round
returns $\theta^\star$ unchanged.
\end{proposition}

\noindent Throughout, $\varepsilon_t=\Delta_t-\alpha_0\Delta_{\mathrm{M}}^0$
is the Stage-0 residual (Eq.~\ref{eq:residual}) and $w$ the headroom
weighting (Eq.~\ref{eq:headroom-weights}).

\paragraph{Step 1: the round-two residuals.}
Taking $\theta^\star$ as the new base, the residual of task $t$ is its task
vector minus everything already applied,
\begin{equation}
\varepsilon'_t=\Delta_t-\big(\alpha_0\Delta_{\mathrm{M}}^0
+\alpha_1^\star\Delta_{\mathrm{M}}^1\big)
=\varepsilon_t-\alpha_1^\star\Delta_{\mathrm{M}}^1 .
\end{equation}

\paragraph{Step 2: the round-two direction.}
With the weights held fixed and $\mathcal{M}(\{x_t\})=\sum_t c_tx_t$,
\begin{align}
\Delta_{\mathrm{M}}^2 &=\mathcal{M}(\{w_t\varepsilon'_t\}) \nonumber\\
&=\sum_t c_tw_t\varepsilon_t-\alpha_1^\star\Big(\sum_t c_tw_t\Big)
   \Delta_{\mathrm{M}}^1 \nonumber\\
&=\Big(1-\alpha_1^\star\sum_t c_tw_t\Big)\,\Delta_{\mathrm{M}}^1 ,
\label{eq:round2dir}
\end{align}
using $\sum_t c_tw_t\varepsilon_t=\mathcal{M}(\{w_t\varepsilon_t\})
=\Delta_{\mathrm{M}}^1$. The round-two direction is therefore a scalar
multiple of the round-one direction. The coefficients $c_t$ must not depend
on the arguments in either step; this, not linearity, is what the
proposition assumes.

\paragraph{Step 3: the reachable set.}
Any model reachable in round two is
\begin{align}
&\theta_0+\alpha_0\Delta_{\mathrm{M}}^0+\alpha_1^\star
  \Delta_{\mathrm{M}}^1+\alpha_2\Delta_{\mathrm{M}}^2 \nonumber\\
&\qquad=\theta_0+\alpha_0\Delta_{\mathrm{M}}^0
  +\alpha_{\mathrm{eff}}\Delta_{\mathrm{M}}^1 ,\nonumber\\
&\alpha_{\mathrm{eff}}=\alpha_1^\star+\alpha_2\Big(1-\alpha_1^\star
  \sum_t c_tw_t\Big).
\label{eq:alphaeff}
\end{align}

\paragraph{Step 4: optimality.}
By Eq.~\ref{eq:alphaeff} the round-two model is the round-one model at
strength $\alpha_{\mathrm{eff}}$. By hypothesis
$\alpha_{\mathrm{eff}}\in\mathcal{G}$ for every $\alpha_2$ considered, and
$\alpha_1^\star$ maximizes that objective over $\mathcal{G}$, so
no non-zero $\alpha_2$ improves on $\alpha_2=0$.
\hfill$\square$

\medskip
\noindent The fixed-coefficient condition holds for weight averaging and task
arithmetic but not for operators whose coefficients depend on their arguments.

\section{Further Related Work}
\label{app:related}

LiNeS~\citep{wang2025lines} rescales a task vector by layer depth after
training; in merging it rescales the merged vector of task arithmetic, TIES,
or consensus merging on the suites we also use. It sets one scale per layer
for all tasks, whereas \proj{} re-merges per-task residuals with headroom
weights; we have neither compared nor combined the two.
AdapterFusion~\citep{pfeiffer2021adapterfusion} trains a fusion layer over
frozen adapters, AdapterSoup~\citep{chronopoulou2023adaptersoup} averages
adapters selected for a new domain, and LoraHub~\citep{huang2024lorahub}
searches LoRA coefficients on a few examples of an unseen task. These methods
adapt to new tasks, whereas \proj{} serves the tasks the adapters were
trained on.

\end{document}